%% file: tog.tex
\documentclass[acmtog]{acmart}

\usepackage{booktabs} % For formal tables
\usepackage{mystyle}
\usepackage{graphicx}
\usepackage{subfig}

\setcitestyle{square}

\usepackage[ruled]{algorithm2e} % For algorithms

\SetAlFnt{\small}
\SetAlCapFnt{\small}
\SetAlCapNameFnt{\small}
\SetAlCapHSkip{0pt}
\IncMargin{-\parindent}

\acmJournal{TOG}
\acmVolume{1}
\acmNumber{1}
\acmArticle{1}
\acmYear{2019}
\acmMonth{1}

\setcopyright{acmlicensed}
\acmDOI{10.1145/3326362}

\begin{document}
% Title portion
\title{Dynamic Graph CNN for Learning on Point Clouds}

% Authors.
\author{Yue Wang}
\affiliation{%
  \institution{Massachusetts Institute of Technology}}
\email{yuewang@csail.mit.edu}

\author{Yongbin Sun}
\affiliation{%
  \institution{Massachusetts Institute of Technology}}
\email{yb_sun@mit.edu}

\author{Ziwei Liu}
\affiliation{%
  \institution{UC Berkeley / ICSI}}
\email{zwliu@icsi.berkeley.edu}

\author{Sanjay E. Sarma}
\affiliation{%
  \institution{Massachusetts Institute of Technology}}
\email{sesarma@mit.edu}

\author{Michael M. Bronstein}
\affiliation{%
  \institution{Imperial College London / USI Lugano}}
\email{m.bronstein@imperial.ac.uk}

\author{Justin M. Solomon}
\affiliation{%
  \institution{Massachusetts Institute of Technology}}
\email{jsolomon@mit.edu}

\newcommand\BigBox{\vcenter{\hbox{\scalebox{2}{$\Box$}}}}
\newcommand\bigsquare{\mathop{\BigBox}\limits}

\input{./sections/abstract.tex}





\begin{teaserfigure}
 \centering
 \includegraphics[width=6.0in]{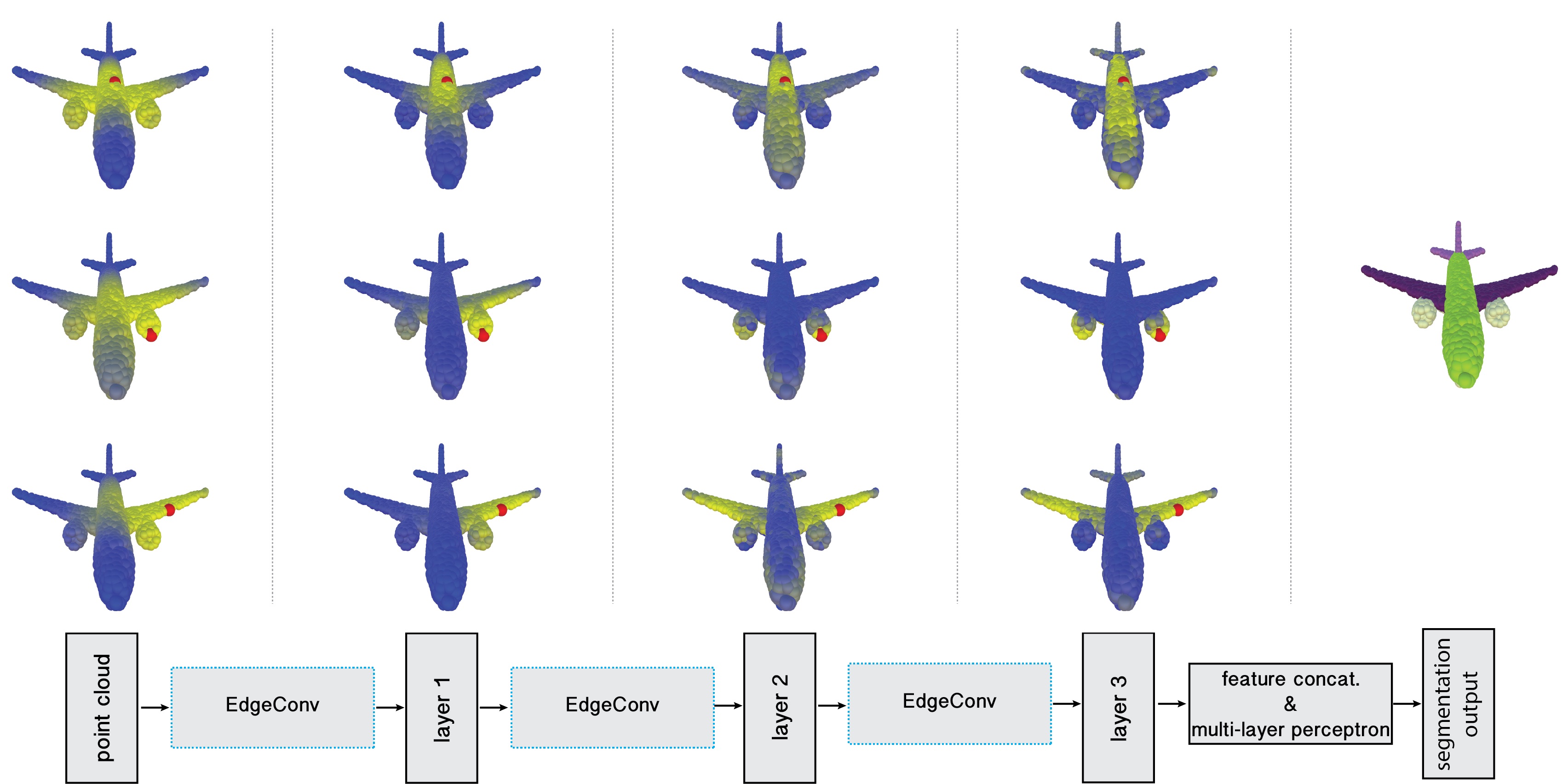}
 \caption{\textbf{Point cloud segmentation using the proposed neural network.} Bottom: schematic neural network architecture. Top: Structure of the feature spaces produced at different layers of the network, visualized as the distance from the red point to all the rest of the points (shown left-to-right are the input and layers 1-3; rightmost figure shows the resulting segmentation). Observe how the feature space structure in deeper layers captures semantically similar structures such as wings, fuselage, or turbines, despite a large distance between them in the original input space.}
\end{teaserfigure}

\maketitle

\input{./sections/introduction.tex}

\input{./sections/related_work.tex}

\input{./sections/approach.tex}
\input{./sections/experiment.tex}

\input{./sections/conclusion.tex}
\input{./sections/acknowledgement.tex}

\bibliographystyle{ACM-Reference-Format}
\bibliography{tog}
\end{document}

%% file: sections/abstract.tex
% !TEX root = ../tog.tex
\begin{abstract}
Point clouds provide a flexible geometric representation suitable for countless applications in computer graphics; they also comprise the raw output of most 3D data acquisition devices. While hand-designed features on point clouds have long been proposed in graphics and vision, however, the recent overwhelming success of convolutional neural networks (CNNs) for image analysis suggests the value of adapting insight from CNN to the point cloud world. Point clouds inherently lack topological information so designing a model to recover topology can enrich the representation power of point clouds. To this end, we propose a new neural network module dubbed \emph{EdgeConv} suitable for CNN-based high-level tasks on point clouds including classification and segmentation. EdgeConv acts on graphs dynamically computed in each layer of the network. It is differentiable and can be plugged into existing architectures.
Compared to existing modules operating in extrinsic space or treating each point independently, EdgeConv has several appealing properties:  It incorporates local neighborhood information; it can be stacked applied to learn global shape properties; and in multi-layer systems affinity in feature space captures semantic characteristics over potentially long distances in the original embedding. We show the performance of our model on standard benchmarks including ModelNet40, ShapeNetPart, and S3DIS.
\end{abstract}

%% file: sections/introduction.tex
% !TEX root = ../tog.tex

\section{Introduction}

Point clouds, or scattered collections of points in 2D or 3D, are arguably the simplest shape representation; they also comprise the output of 3D sensing technology including LiDAR scanners and stereo reconstruction.  With the advent of fast 3D point cloud acquisition, recent pipelines for graphics and vision often process point clouds directly, bypassing expensive mesh reconstruction or denoising due to efficiency considerations or instability of these techniques in the presence of noise.  A few of the many recent applications of point cloud processing and analysis include indoor navigation \cite{zhu2017}, self-driving vehicles \cite{qi2017frustum,wang2018cvpr,liang2018eccv}, robotics \cite{Rusu08RAS2}, and shape synthesis and modeling \cite{golovinskiy09,pcpnet}. 

These modern applications demand \emph{high-level} processing of point clouds.  Rather than identifying salient geometric features like corners and edges, recent algorithms search for semantic cues and affordances.  These features do not fit cleanly into the frameworks of computational or differential geometry and typically require learning-based approaches that derive relevant information through statistical analysis of labeled or unlabeled datasets.  

In this paper, we primarily consider point cloud classification and segmentation, two model tasks in point cloud processing.  Traditional methods for solving these problems employ handcrafted features to capture geometric properties of point clouds \cite{lu2014recognizing, rusu2009fast, rusu2008aligning}. More recently, the success of deep neural networks for image processing has motivated a data-driven approach to learning features on point clouds.  Deep point cloud processing and analysis methods are developing rapidly and outperform traditional approaches in various tasks \cite{chang2015shapenet}.

Adaptation of deep learning to point cloud data, however, is far from straightforward.  Most critically, standard deep neural network models require input data with regular structure, while point clouds are fundamentally irregular: Point positions are continuously distributed in the space, and any permutation of their ordering does not change the spatial distribution. One common approach to process point cloud data using deep learning models is to first convert raw point cloud data into a volumetric representation, namely a 3D grid \cite{maturana2015voxnet, wu20153d}. This approach, however, usually introduces quantization artifacts and excessive memory usage, making it difficult to go to capture high-resolution or fine-grained features.

State-of-the-art deep neural networks are designed specifically to handle the irregularity of point clouds, directly manipulating raw point cloud data rather than passing to an intermediate regular representation.  This approach was pioneered by \textit{PointNet} \cite{DBLP:journals/corr/QiSMG16}, which achieves permutation invariance of points by operating on each point independently and subsequently applying a symmetric function to accumulate features. Various extensions of PointNet consider neighborhoods of points rather than acting on each independently \cite{DBLP:journals/corr/QiYSG17, shen2017neighbors}; these allow the network to exploit local features, improving upon performance of the basic model. These techniques largely treat points independently at local scale to maintain permutation invariance. This independence, however, neglects the geometric relationships among points, presenting a fundamental limitation that cannot capture local features. 

To address these drawbacks, we propose a novel simple operation, called EdgeConv, which captures local geometric structure while maintaining permutation invariance. Instead of generating point features directly from their embeddings, EdgeConv generates \textit{edge features} that describe the relationships between a point and its neighbors. EdgeConv is designed to be invariant to the ordering of neighbors, and thus is permutation invariant. Because EdgeConv explicitly constructs a local graph and learns the embeddings for the edges, the model is capable of grouping points both in Euclidean space and in semantic space.

EdgeConv is easy to implement and integrate into existing deep learning models to improve their performance. In our experiments, we integrate EdgeConv into the basic version of \textit{PointNet} without using any feature transformation. We show %performance improvement by a large margin; 
the resulting network achieves state-of-the-art performance on several datasets, most notably \textit{ModelNet40} and \textit{S3DIS} for classification and segmentation.

\paragraph*{Key Contributions.} We summarize the key contributions of our work as follows:
  \begin{itemize}[leftmargin=*]
  \item  We present a novel operation for learning from point clouds, EdgeConv, to better capture local geometric features of point clouds while still maintaining permutation invariance.
  \item  We show the model can learn to semantically group points by dynamically updating a graph of relationships from layer to layer.
  \item  We demonstrate that EdgeConv can be integrated into multiple existing pipelines for point cloud processing.
  \item  We present extensive analysis and testing of EdgeConv and show that it achieves state-of-the-art performance on benchmark datasets.
  \item We release our code to facilitate reproducibility and future research. \footnote{\url{https://github.com/WangYueFt/dgcnn}}
\end{itemize}

\begin{figure*}[t!]
  \centering
  \includegraphics[width=0.9\textwidth]{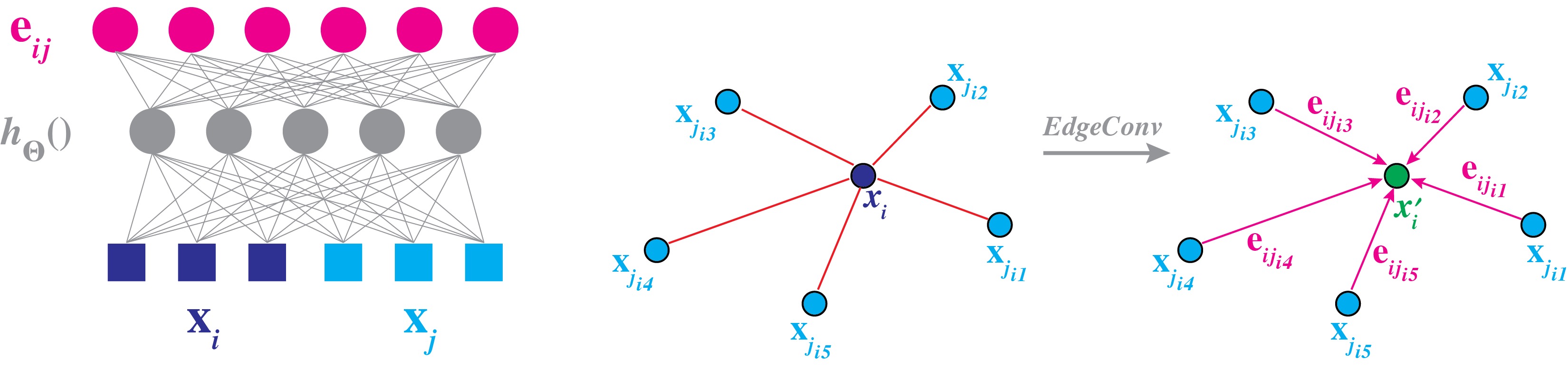}
  \caption{\textbf{Left}: Computing an edge feature, $\e_{ij}$ (top), from a point pair, $\x_i$ and $\x_j$ (bottom). In this example, $h_{\boldsymbol{\Theta}}()$ is instantiated using a fully connected layer, and the learnable parameters are its associated weights.  \textbf{Right}: The EdgeConv operation. The output of EdgeConv is calculated by aggregating the edge features associated with all the edges emanating from each connected vertex.}
  \label{fig:edgeconv}
\end{figure*}

%% file: sections/related_work.tex
% !TEX root = ../tog.tex
\section{Related Work}

\paragraph*{Hand-Crafted Features} Various tasks in geometric data processing and analysis---including segmentation, classification, and matching---require some notion of local similarity between shapes. Traditionally, this similarity is established by constructing feature descriptors that capture local geometric structure. Countless papers in computer vision and graphics propose local feature descriptors for point clouds suitable for different problems and data structures.  A comprehensive overview of hand-designed point features is out of the scope of this paper, but we refer the reader to \cite{van2011survey, guo20143d, biasotti2016recent} for discussion.

Broadly speaking, one can distinguish between \textit{extrinsic} and \textit{intrinsic} descriptors. Extrinsic descriptors usually are derived from the coordinates of the shape in 3D space and includes classical methods like shape context \cite{belongie2001shape}, spin images \cite{johnson1999using}, integral features \cite{manay2006integral}, distance-based descriptors \cite{ling2007shape}, point feature histograms \cite{rusu2008aligning, rusu2009fast}, and normal histograms \cite{tombari2011combined}, to name a few. Intrinsic descriptors treat the 3D shape as a manifold whose metric structure is discretized as a mesh or graph; quantities expressed in terms of the metric are invariant to isometric deformation. Representatives of this class include spectral descriptors such as global point signatures \cite{rustamov2007laplace}, the heat and wave kernel signatures \cite{sun2009concise,aubry2011wave}, and variants \cite{bronstein2010scale}. Most recently, several approaches wrap machine learning schemes around standard descriptors \cite{guo20143d, shah20133d}. 

\paragraph*{Deep learning on geometry} Following the breakthrough results of convolutional neural networks (CNNs) in vision \cite{lecun1989backpropagation,krizhevsky2012imagenet}, there has been strong interest to adapt such methods to geometric data. 
Unlike images, geometry usually does not have an underlying grid, requiring new building blocks replacing convolution and pooling or adaptation to a grid structure. 

As a simple way to overcome this issue, view-based \cite{su2015multi,wei2016dense} and volumetric representations \cite{maturana2015voxnet,wu20153d,klokov2017escape,tatarchenko2017octree}---or their combination \cite{qi2016volumetric}---``place'' geometric data onto a grid.
More recently, PointNet~\cite{DBLP:journals/corr/QiSMG16, DBLP:journals/corr/QiYSG17} exemplifies a broad class of deep learning architectures on non-Euclidean data (graphs and manifolds) termed 
\textit{geometric deep learning} \cite{bronstein2017geometric}. These date back to early methods to construct neural networks on graphs \cite{scarselli2009graph}, recently improved with gated recurrent units \cite{li2016gated} and neural message passing \cite{gilmer2017neural}.
\citet{bruna2013spectral} and \citet{henaff2015deep} generalized convolution to graphs via the Laplacian eigenvectors
\cite{shuman2013emerging}. 
Computational drawbacks of this foundational approach were alleviated in follow-up works using polynomial \cite{defferrard2016convolutional,kipf2016semi, monti2017geometric,monti2018motifnet}, or rational \cite{levie2017cayleynets} spectral filters that avoid Laplacian eigendecomposition and guarantee localization. 
An alternative definition of non-Euclidean convolution employs spatial rather than spectral filters. The \textit{Geodesic CNN (GCNN)} is a deep CNN on meshes generalizing the notion of patches using local intrinsic parameterization \cite{masci2015geodesic}. Its key advantage over spectral approaches is better generalization as well as a simple way of constructing directional filters. Follow-up work proposed different local charting techniques using anisotropic diffusion \cite{boscaini2016learning} or Gaussian mixture models \cite{velivckovic2017graph, monti2016geometric}.  In \cite{litany2017deep, halimi2018self}, a differentiable functional map \cite{ovsjanikov2012functional} layer was incorporated into a geometric deep neural network, allowing to do intrinsic structured prediction of correspondence between nonrigid shapes.

The last class of geometric deep learning approaches attempts to pull back a convolution operation by embedding the shape into a domain with shift-invariant structure such as the sphere \cite{sinha2016deep}, torus \cite{maron2017convolutional}, plane \cite{ezuz2017gwcnn}, sparse network lattice \cite{su18splatnet}, or spline \cite{splinecnn2018}.

Finally, we should mention {\em geometric generative models}, which attempt to generalize models such as autoencoders, variational autoencoders (VAE) \cite{kingma2013auto}, and generative adversarial networks (GAN) \cite{goodfellow2014generative} to the non-Euclidean setting. One of the fundamental differences between these two settings is the lack of canonical order between the input and the output vertices, thus requiring an input-output correspondence problem to be solved. In 3D mesh generation, it is commonly assumed that the mesh is given and its vertices are canonically ordered; the generation problem thus amounts only to determining the embedding of the mesh vertices. \citet{kostrikov2017surface} proposed SurfaceNets based on the extrinsic Dirac operator for this task. \citet{litany2017deformable} introduced the intrinsic VAE for meshes and applied it to shape completion; a similar architecture was used by \citet{ranjan2018generating} for 3D face synthesis. For point clouds, multiple generative architectures have been proposed \cite{fan2017point, li2018point, foldingnet2018}. 

\begin{figure*}[t!]
  \centering
     \includegraphics[width=1.0\textwidth]{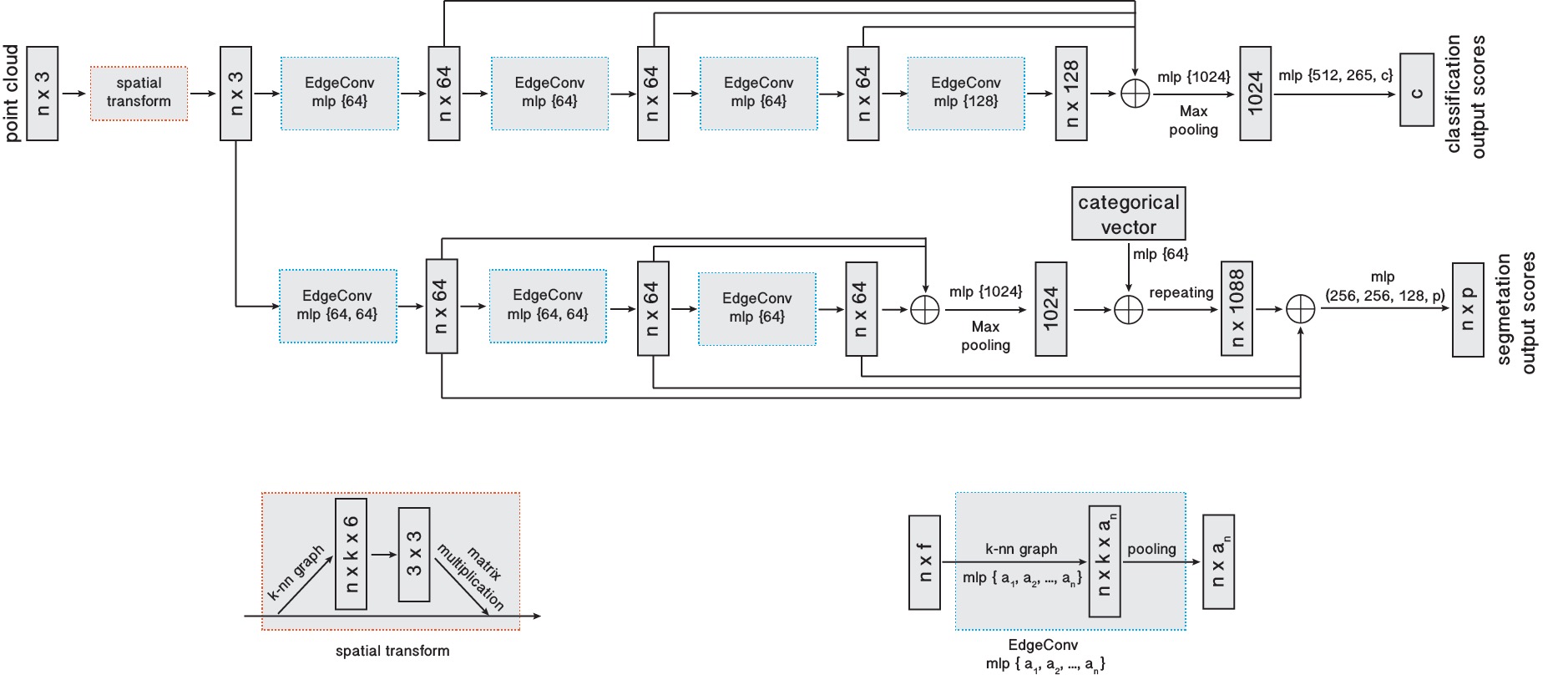}
  \caption{\textbf{Model architectures:} The model architectures used for classification (top branch) and segmentation (bottom branch). The classification model takes  as input $n$ points, calculates an edge feature set of size $k$ for each point at an EdgeConv layer, and aggregates features within each set to compute EdgeConv responses for corresponding points. The output features of the last EdgeConv layer are aggregated globally to form an $1D$ global descriptor, which is used to generate classification scores for $c$ classes. The segmentation model extends the classification model by concatenating the $1D$ global descriptor and all the EdgeConv outputs (serving as local descriptors) for each point. It outputs per-point classification scores for $p$ semantic labels. 
  $\oplus$: concatenation.
  \textbf{Point cloud transform block:} The point cloud transform block is designed to align an input point set to a canonical space by applying an estimated $3\times3$ matrix. To estimate the $3\times3$ matrix, a tensor concatenating the coordinates of each point and the coordinate differences between its $k$ neighboring points is used. \textbf{EdgeConv block:} The EdgeConv block takes as input a tensor of shape $n\times f$, computes edge features for each point by applying a multi-layer perceptron (mlp) with the number of layer neurons defined as $\{a_1, a_2, ..., a_n\}$, and generates a tensor of shape $n\times a_n$ after pooling among neighboring edge features.}
  \label{fig:model_architecture}
\end{figure*}

%% file: sections/approach.tex
% !TEX root = ../tog.tex
\section{Our approach}

We propose an approach inspired by PointNet and convolution operations. Instead of working on individual points like PointNet, however, we exploit local geometric structures by constructing a local neighborhood graph and applying convolution-like operations on the edges connecting neighboring pairs of points, in the spirit of graph neural networks. We show in the following that such an operation, dubbed \textit{edge convolution} (EdgeConv), has properties lying between translation-invariance and non-locality. 

Unlike graph CNNs, our graph is not fixed but rather is dynamically updated after each layer of the network. That is, the set of $k$-nearest neighbors of a point changes from layer to layer of the network and is computed from the sequence of embeddings. Proximity in feature space differs from proximity in the input, leading to nonlocal diffusion of information throughout the point cloud. As a connection to existing work, Non-local Neural Networks \cite{NonLocal2018} explored similar ideas in the video recognition field, and follow-up work by \citet{xie2018feature} proposed using non-local blocks to denoise feature maps to defend against adversarial attacks.

\subsection{Edge Convolution} 
Consider an $F$-dimensional point cloud with $n$ points, denoted by $\mathbf{X} = \{\mathbf{x}_1, \hdots, \mathbf{x}_n\} \subseteq \mathbb R^F$. 
In the simplest setting of $F=3$, each point contains 3D coordinates $\mathbf{x}_i = (x_i, y_i, z_i)$; it is also possible to include additional coordinates representing color, surface normal, and so on. 
In a deep neural network architecture, each subsequent layer operates on the output of the previous layer, so more generally the dimension $F$ represents the feature dimensionality of a given layer. 

We compute a directed graph $\mathcal{G} = ( \mathcal{V}, \mathcal{E} )$ representing local point cloud structure, where $\mathcal{V} = \{1, \hdots, n\}$ and $\mathcal{E} \subseteq \mathcal{V} \times \mathcal{V}$ are the \textit{vertices} and \textit{edges}, respectively.
In the simplest case, we construct $\mathcal{G}$ as the $k$-nearest neighbor ($k$-NN) graph of $\mathbf{X}$ in $\mathbb{R}^F$. The graph includes self-loop, meaning each node also points to itself.
We define \textit{edge features} as $\boldsymbol{e}_{ij} = h_{\boldsymbol{\Theta}}(\mathbf{x}_i, \mathbf{x}_j)$, where $h_{\boldsymbol{\Theta}} : \mathbb{R}^F\times \mathbb{R}^F \rightarrow \mathbb{R}^{F'}$ is a nonlinear function with a set of learnable  parameters $\boldsymbol{\Theta}$. 
 
Finally, we define the EdgeConv operation by applying a channel-wise symmetric aggregation operation $\square$ (e.g., $\sum$ or $\max$) on the edge features associated with all the edges emanating from each vertex. The output of EdgeConv at the $i$-th vertex is thus given by %
\begin{equation}
\label{eq:full}
\mathbf{x}'_i  =\mathop{\bigsquare}_{j: (i,j) \in \mathcal{E}} h_{\boldsymbol{\Theta}}(\mathbf{x}_i, \mathbf{x}_j).
\end{equation}
Making analogy to convolution along images, we regard $\mathbf{x}_i$ as the central pixel and $\{ \mathbf{x}_j : (i,j) \in \mathcal{E}\}$ as a patch around it (see Figure~\ref{fig:edgeconv}).  
Overall, given an $F$-dimensional point cloud with $n$ points, EdgeConv produces an $F'$-dimensional point cloud with the same number of points. 

\textit{Choice of $h$ and $\square$.}
The choice of the edge function and the aggregation operation has a crucial influence on the properties of EdgeConv.
For example, when $\mathbf{x}_1, \hdots, \mathbf{x}_n$ represent image pixels on a regular grid and the graph $\mathcal G$ has connectivity representing patches of fixed size around each pixel, the choice $\boldsymbol{\theta_m} \cdot \mathbf{x}_j$ as the edge function and sum as the aggregation operation yields standard convolution:
\begin{equation}
\label{eq:choice1}
x'_{im} = \sum_{j: (i,j) \in \mathcal{E}} \boldsymbol{\theta}_m\cdot\mathbf{x}_j,
\end{equation}
Here, $\boldsymbol{\Theta} = (\boldsymbol{\theta}_1, \hdots, \boldsymbol{\theta}_M)$ encodes the weights of $M$ different filters. Each $\boldsymbol{\theta}_m$ has the same dimensionality as $\x$, and $\cdot$ denotes the Euclidean inner product. 

A second  choice of $h$ is 
\begin{equation}
    \label{eq:choice2}
    h_{\boldsymbol{\Theta}}(\mathbf{x}_i, \mathbf{x}_j) = h_{\boldsymbol{\Theta}}(\mathbf{x}_i),
\end{equation}
encoding only global shape information oblivious of the local neighborhood structure. This type of operation is used in PointNet, which can thus be regarded as a special case of EdgeConv. 

A third choice of $h$ adopted by \citet{pcnn2018} is 
\begin{equation}
    \label{eq:choice31}
    h_{\boldsymbol{\Theta}}(\mathbf{x}_i,
    \mathbf{x}_j) = h_{\boldsymbol{\Theta}}(
    \mathbf{x}_j)
\end{equation}
and
\begin{equation}
    \label{eq:choice32}
     x'_{im}=\sum_{j \in \mathcal{V}} (h_{\boldsymbol{\theta}(
    \mathbf{x}_j)}) g(u(\mathbf{x}_i, \mathbf{x}_j)),
\end{equation}
where $g$ is a Gaussian kernel and $u$ computes pairwise distance in Euclidean space. 

A fourth option is 
\begin{equation}
    \label{eq:choice4}
    h_{\boldsymbol{\Theta}}(\mathbf{x}_i, \mathbf{x}_j) = h_{\boldsymbol{\Theta}}(\mathbf{x}_j - \mathbf{x}_i).
\end{equation}
This encodes only local information, treating the shape as a collection of small patches and losing global structure. 

Finally, a fifth option that we adopt in this paper is an asymmetric edge function 
\begin{equation}
\label{eq:choice51}
h_{\boldsymbol{\Theta}}(\mathbf{x}_i, \mathbf{x}_j) = \bar h_{\boldsymbol{\Theta}}(\mathbf{x}_i, \mathbf{x}_j - \mathbf{x}_i).
\end{equation}
This explicitly combines global shape structure, captured by the coordinates of the patch centers $\mathbf{x}_i$, with local neighborhood information, captured by $\mathbf{x}_j - \mathbf{x}_i$. In particular, we can define our operator by notating
 \begin{equation}
 \label{eq:choice52}
    e_{ijm}' = \mathrm{ReLU}(\boldsymbol{\theta}_m \cdot (\mathbf{x}_j-\mathbf{x}_i) + \boldsymbol{\phi}_m \cdot \mathbf{x}_i),
\end{equation}
which can be implemented as a shared MLP, and taking
\begin{equation}
    \label{eq:choice53}
    x'_{im} = \max_{j: (i,j) \in \mathcal{E}}e_{ijm}',
\end{equation}
where $\boldsymbol{\Theta} = (\boldsymbol{\theta}_1, \hdots, \boldsymbol{\theta}_M, \boldsymbol{\phi}_1, \hdots, \boldsymbol{\phi}_M)$

\subsection{Dynamic graph update} 
Our experiments suggest that it is beneficial to \textit{recompute the graph} using nearest neighbors in the feature space produced by each layer. This is a crucial distinction of our method from graph CNNs working on a fixed input graph. Such a dynamic graph update is the reason for the name of our architecture, the \textit{Dynamic Graph CNN (DGCNN)}. With dynamic graph updates, the receptive field is as large as the diameter of the point cloud, while being sparse.

At each layer we have a different graph 
$\mathcal{G}^{(l)} = (\mathcal{V}^{(l)}, \mathcal{E}^{(l)})$, where the $l$-th layer  edges are of the form $(i, j_{i1}), \hdots, (i,j_{ik_l})$ such that $\x^{(l)}_{j_{i1}}, \hdots, x^{(l)}_{j_{ik_l}}$ are the $k_l$ points closest to $\mathbf{x}^{(l)}_i$. 
Put differently, our architecture learns \emph{how} to construct the graph $\mathcal G$ used in each layer rather than taking it as a fixed constant constructed before the network is evaluated. In our implementation, we compute a pairwise distance matrix in feature space and then take the closest $k$ points for each single point.

\begin{table*}[t!]
\vskip 0.1in
\begin{center}
\resizebox{2\columnwidth}{!}{
\begin{tabular}{lcccr}
\toprule
   & Aggregation & Edge Function & Learnable parameters \\
\midrule
PointNet \cite{DBLP:journals/corr/QiSMG16} & --- & $h_{\boldsymbol{\Theta}}(\mathbf{x}_i, \mathbf{x}_j) = h_{\boldsymbol{\Theta}}(\mathbf{x}_i)$ & $\boldsymbol{\Theta}$\\
PointNet++ \cite{DBLP:journals/corr/QiYSG17} & $\max$ & $h_{\boldsymbol{\Theta}}(\mathbf{x}_i, \mathbf{x}_j) = h_{\boldsymbol{\Theta}}(\mathbf{x}_j)$ & $\boldsymbol{\Theta}$ \\
MoNet \cite{monti2016geometric} & $\sum$ & $h_{\boldsymbol{\theta}_m, \boldsymbol{w_n}}(\mathbf{x}_i, \mathbf{x}_j) = \boldsymbol{\theta}_m\cdot (\mathbf{x}_j \odot g_{\boldsymbol{w}_n}(u(\mathbf{x}_i, \mathbf{x}_j)))$ & $\boldsymbol{w_n},  \boldsymbol{\theta}_m$\\

PCNN \cite{pcnn2018} & $\sum$ 
& $h_{\boldsymbol{\theta}_m}(\mathbf{x}_i,\mathbf{x}_j)=
 (\boldsymbol{\theta}_m\cdot \mathbf{x}_j) g(u(\mathbf{x}_i, \mathbf{x}_j))
$
& $\boldsymbol{\theta_m}$ \\

\bottomrule
\end{tabular}
}
\end{center}
\caption{Comparison to existing methods. The per-point weight $w_i$ in \cite{pcnn2018} effectively is computed in the first layer and could be carried onward as an extra feature; we omit this for simplicity.}
\label{table:sum}
\end{table*}

\begin{figure*}[t!]
  \centering
 \includegraphics[width=1.0\textwidth]{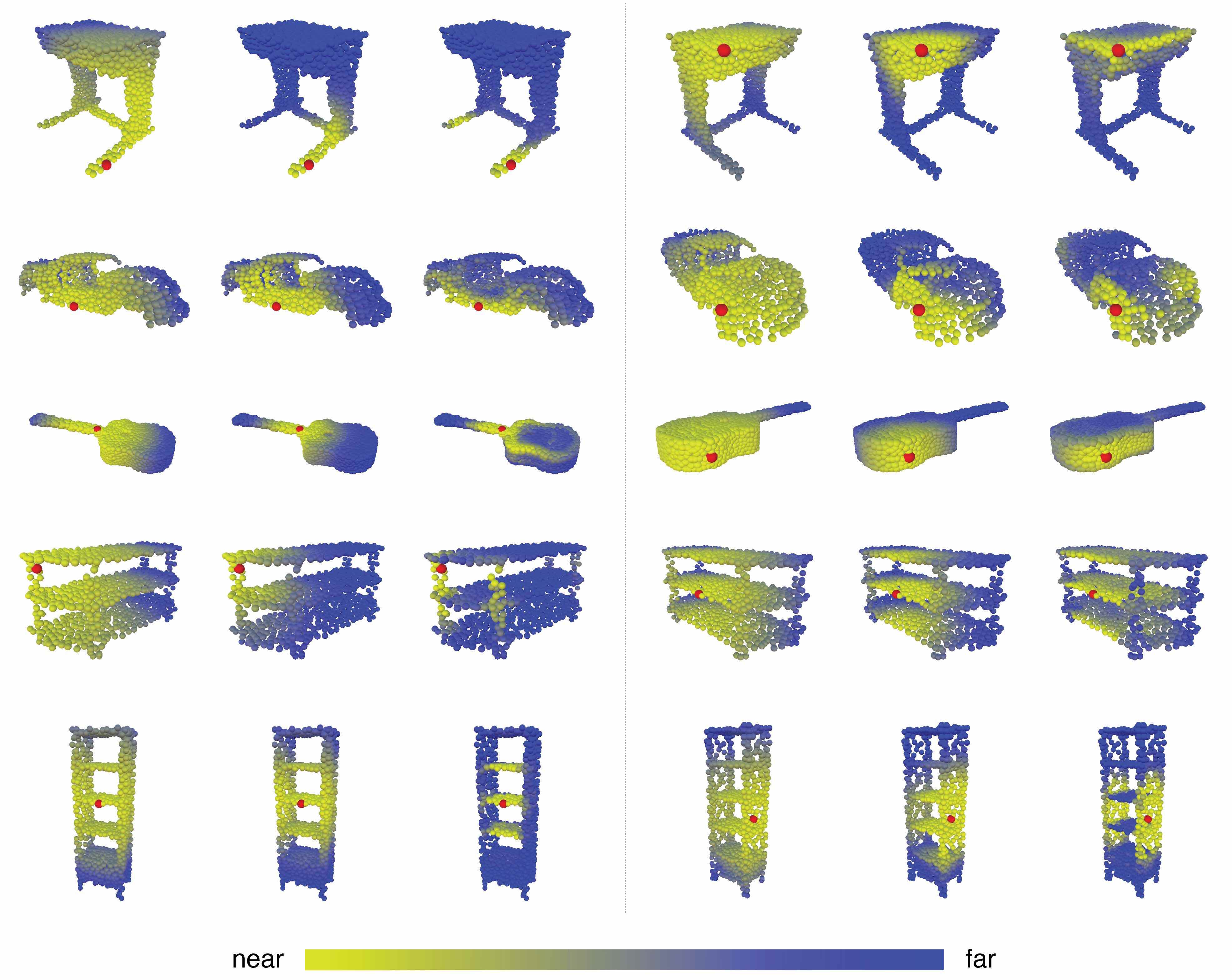}
  \caption{{\bf Structure of the feature spaces} produced at different stages of our shape classification neural network architecture, visualized as the distance between the red point to the rest of the points. For each set, \textbf{Left:} Euclidean distance in the input $\mathbb{R}^3$ space; \textbf{Middle:} Distance after the point cloud transform stage, amounting to a global transformation of the shape; \textbf{Right:} Distance in the feature space of the last layer. Observe how in the feature space of deeper layers semantically similar structures such as shelves of a bookshelf or legs of a table are brought close together, although they are distant in the original space.}
  \label{fig:feature_closeness}
\end{figure*}

\subsection{Properties}

\paragraph*{Permutation Invariance.}
Consider the output of a layer 
\begin{equation}
\label{eq:permutation}
    \x'_{i}=\mathop{\max}_{j: (i,j) \in \mathcal{E}} h_{\boldsymbol{\Theta}}(\x_i, \x_j)
\end{equation}
and a permutation operator $\pi$. The output of the layer $\x'_{i}$ is invariant to permutation of the input $\x_j$ because $\max$ is a symmetric function (other symmetric functions also apply). The global max pooling operator to aggregate point features is also permutation-invariant.

\paragraph*{Translation Invariance.}
Our operator has a ``partial'' translation invariance property, in that our choice of edge functions~\eqref{eq:choice51} explicitly exposes the part of the function that can be translation-dependent and optionally can be disabled. Consider a translation  applied to $\x_j$ and $\x_i$; we can show that part of the edge feature is preserved when shifting by $\boldsymbol{T}$. In particular, for the translated point cloud we have 
\begin{align*}
\label{eq:translation}
e_{ijm}' 
&= \boldsymbol{\theta}_m\cdot(\mathbf{x}_j+T-(\mathbf{x}_i+T)) + \boldsymbol{\phi}_m\cdot(\mathbf{x}_i + T)\\
&= \boldsymbol{\theta}_m\cdot(\mathbf{x}_j-\mathbf{x}_i) + \boldsymbol{\phi}_m\cdot(\mathbf{x}_i + T).
\end{align*}

If we only consider $\x_j-\x_i$ by taking $\boldsymbol{\phi}_m=\boldsymbol{0}$, then the operator is fully invariant to translation. In this case, however, the model reduces to recognizing an object based on an unordered set of patches, ignoring the positions and orientations of patches. With both $\x_j-\x_i$ and $\x_i$ as input, the model takes account into the local geometry of patches while keeping global shape information. 

\subsection{Comparison to existing methods} 

DGCNN is related to two classes of approaches, PointNet and graph CNNs, which we show to be particular settings of our method. We summarize different methods in Table~\ref{table:sum}.

PointNet is a special case of our method with $k=1$, yielding a graph with an empty edge set $\mathcal{E} = \varnothing$. The edge function used in PointNet is $h_{\boldsymbol{\Theta}}(\mathbf{x}_i, \mathbf{x}_j) = h_{\boldsymbol{\Theta}}(\mathbf{x}_i)$, which considers global but not local geometry. 
PointNet++ tries to account for local structure by applying PointNet in a local manner. In our parlance, PointNet++ first constructs the graph according to the Euclidean distances between the points, and in each layer applies a graph coarsening operation. For each layer, some points are selected using farthest point sampling (FPS); only the selected points are preserved while others are directly discarded after this layer. In this way, the graph becomes smaller after the operation applied on each layer. In contrast to DGCNN, PointNet++ computes pairwise distances using point input coordinates, and hence their graphs are fixed during training.  The edge function used by PointNet++ is $h_{\boldsymbol{\Theta}}(\mathbf{x}_i, \mathbf{x}_j) = h_{\boldsymbol{\Theta}}(\mathbf{x}_j)$, and the aggregation operation is also a $\max$. 

Among graph CNNs, MoNet \cite{monti2016geometric}, ECC \cite{simonovsky2017ecc}, Graph Attention Networks \cite{velivckovic2017graph}, and the concurrent work \cite{pcnn2018} are the most related approaches. Their common denominator is a notion of a local patch on a graph, in which a convolution-type operation can be defined.\footnote{\cite{simonovsky2017ecc,velivckovic2017graph} can be considered instances of \cite{monti2016geometric}, with the difference that the weights are constructed employing features from adjacent nodes instead of graph structure; \cite{pcnn2018} is also similar except that the weighting function is hand-designed.}

Specifically, \citet{monti2016geometric} use the graph structure to compute a local ``pseudo-coordinate system'' $\mathbf{u}$ in which the neighborhood vertices are represented; the convolution is then defined as an $M$-component Gaussian mixture
\begin{equation}
\label{eq:monet}
     x'_{im}=\sum_{j: (i, j) \in \mathcal{E}} \boldsymbol{\theta}_m \cdot (\mathbf{x}_j \odot g_{\boldsymbol{w_n}}(u(\mathbf{x}_i, \mathbf{x}_j))),
\end{equation}
where $g$ is a Gaussian kernel, $\odot$ is the elementwise (Hadamard) product,  $\{\boldsymbol{w}_1, \hdots, \boldsymbol{w}_N\}$ encode the learnable parameters of the Gaussians (mean and covariance), and $\{\theta_1, \hdots, \theta_M\}$ are the learnable filter coefficients. 
\eqref{eq:monet} is an instance of our general operation~\eqref{eq:full}, with a particular edge function $$h_{ \boldsymbol{\theta}_m, \boldsymbol{w_n}}(\mathbf{x}_i, \mathbf{x}_j) = \boldsymbol{\theta}_m\cdot (\mathbf{x}_j \odot g_{\boldsymbol{w}_n}(u(\mathbf{x}_i, \mathbf{x}_j)))$$ and $\square = \sum$. Again, their graph structure is fixed, and $u$ is constructed based on the degrees of nodes. 

\cite{pcnn2018} can be seen as a special case of \cite{monti2016geometric} with $g$ as predefined Gaussian functions. Removing learnable parameters $(w_1, \hdots, w_N)$ and constructing a dense graph from point clouds, we have 
\begin{equation}
\label{eq:pcnn}
x'_{im}=\sum_{j: j \in \mathcal{V}} (\boldsymbol{\theta}_m\cdot \mathbf{x}_j) g(u(\mathbf{x}_i, \mathbf{x}_j)),
\end{equation}
where $u$ is the pairwise distance between $\x_i$ and $\x_j$ in Euclidean space. 

While MoNet and other graph CNNs assume a given fixed graph on which convolution-like operations are applied, to our knowledge our method is the first for which the graph changes from layer to layer and even on the same input during training when learnable parameters are updated. 
This way, our model not only learns how to extract local geometric features, but also how to group points in a point cloud.  Figure~\ref{fig:feature_closeness} shows the distance in different feature spaces, exemplifying that the distances in deeper layers carry semantic information over long distances in the original embedding.

%% file: sections/experiment.tex
% !TEX root = ../tog.tex

\section{Evaluation}

In this section, we evaluate the models constructed using EdgeConv for different tasks: classification, part segmentation, and semantic segmentation. We also visualize experimental results to illustrate key differences from previous work. 

\subsection{Classification}
\label{sec:classification}
\paragraph{Data} We evaluate our model on the ModelNet40~\cite{wu20153d} classification task, consisting in predicting the category of a previously unseen shape. The dataset contains 12,311 meshed CAD models from 40 categories. 9,843 models are used for training and 2,468 models are for testing. We follow verbatim the experimental settings of  \citet{DBLP:journals/corr/QiSMG16}. For each model, 1,024 points are uniformly sampled from the mesh faces; the point cloud is rescaled to fit into the unit sphere. Only the $(x,y,z)$ coordinates of the sampled points are used, and the original meshes are discarded. During the training procedure, we augment the data by randomly scaling objects and perturbing the object and point locations.

\paragraph{Architecture}
The network architecture used for the classification task is shown in Figure~\ref{fig:model_architecture} (top branch without spatial transformer network). 
We use four EdgeConv layers to extract geometric features. The four EdgeConv layers use three shared fully-connected layers $(64$, $64$, $128$, $256)$. We recompute the graph based on the features of each EdgeConv layer and use the new graph for next layer. The number $k$ of nearest neighbors is 20 for all EdgeConv layers (for the last row in Table \ref{table:cls}, $k$ is 40).
Shortcut connections are included to extract multi-scale features and one shared fully-connected layer $(1024)$ to aggregate multi-scale features, where we concatenate features from previous layers to get a 64+64+128+256=512 dimensional point cloud. Then, a global max/sum pooling is used to get the point cloud global feature, after which two fully-connected layers $(512, 256)$ are used to transform the global feature. Dropout with keep probability of 0.5 is used in the last two fully-connected layers. All layers include LeakyReLU and batch normalization. The number $k$ was chosen using a validation set. We split the training data to 80\% for training and 20\% for validation to search the best $k$. After $k$ is chosen, we retrain the model on the whole training data and evaluate the model on the testing data. Other hyperparameters were chosen in a similar ways.

\paragraph{Training}
We use SGD with learning rate 0.1, and we reduce the learning rate until 0.001 using cosine annealing~\cite{sgdr}. The momentum for batch normalization is 0.9, and we do not use batch normalization decay. The batch size is 32 and the momentum is 0.9.

\paragraph{Results}
Table~\ref{table:cls} shows the results for the classification task. Our model achieves the best results on this dataset. Our baseline using a fixed graph determined by proximity in the input point cloud is $1.0\%$ better than PointNet++. An advanced version including dynamical graph recomputation achieves the best results on this dataset. All the experiments are performed with point clouds that contain 1024 points except last row. We further test out model with 2048 points. The $k$ used for 2048 points is 40 to maintain the same density. Note that PCNN \cite{pcnn2018} uses additional augmentation techniques like randomly sampling 1024 points out of 1200 points during both training and testing.

\begin{table}[H]
\vskip 0.1in
\begin{center}
\resizebox{\columnwidth}{!}{
\begin{small}
\begin{sc}
\begin{tabular}{lccr}
\toprule
   & Mean & Overall  \\
   &  Class Accuracy & Accuracy \\
\midrule
3DShapeNets \cite{wu20153d} & 77.3 & 84.7 \\
VoxNet \cite{maturana2015voxnet} & 83.0 & 85.9 \\
Subvolume \cite{qi2016volumetric} & 86.0 & 89.2 \\
VRN (single view) \cite{brock2016} & 88.98 & - \\
VRN (multiple views) \cite{brock2016} & 91.33 & - \\
ECC \cite{simonovsky2017ecc} & 83.2 & 87.4 \\
PointNet \cite{DBLP:journals/corr/QiSMG16}                   & 86.0 & 89.2 \\
PointNet++ \cite{DBLP:journals/corr/QiYSG17}                    & - & 90.7  \\
Kd-net \cite{klokov2017escape}
    & - & 90.6  \\
PointCNN \cite{pointcnn}
    & 88.1 & 92.2 \\
PCNN \cite{pcnn2018}
    & - & 92.3  \\
\midrule
Ours (baseline)                         & 88.9 & 91.7 \\
Ours                         & \textbf{90.2} & \textbf{92.9} \\
Ours (2048 points)                 & \textbf{90.7} & \textbf{93.5} \\
\bottomrule
\end{tabular}
\end{sc}
\end{small}
}
\end{center}
% \vskip -0.1in
\caption{Classification results on ModelNet40.}
\label{table:cls}
\end{table}

\subsection{Model Complexity}
\label{sec:model_complexity}
We use the ModelNet40~\cite{wu20153d} classification experiment to compare the complexity of our model to previous state-of-the-art. Table~\ref{table:cls_time} shows that our model achieves the best tradeoff between the model complexity (number of parameters), computational complexity (measured as forward pass time), and the resulting classification accuracy. 

Our baseline model using the fixed $k$-NN graph outperforms the previous state-of-the-art PointNet++ by $1.0\%$ accuracy, at the same time being 7 times faster. 
A more advanced version of our model including a dynamically-updated graph computation outperforms PointNet++, PCNN by $2.2\%$ and $0.6\%$ respectively, while being much more efficient. The number of points in each experiment is also 1024 in this section.

\begin{table}[t]
\vskip 0.1in
\begin{center}
\resizebox{\columnwidth}{!}{
\begin{small}
\begin{sc}
\begin{tabular}{lcccr}
\toprule
   & Model size(MB) & Time(ms)  & Accuracy(\%) \\
\midrule
PointNet (Baseline) \cite{DBLP:journals/corr/QiSMG16} & 9.4 & 6.8 & 87.1 \\
PointNet \cite{DBLP:journals/corr/QiSMG16} & 40 & 16.6 & 89.2 \\
PointNet++ \cite{DBLP:journals/corr/QiYSG17} & 12 & 163.2 & 90.7\\
PCNN \cite{pcnn2018} & 94 & 117.0 & 92.3 \\
Ours (Baseline) & 11 & 19.7 & 91.7\\
Ours & 21 & 27.2 & 92.9 \\

\bottomrule
\end{tabular}
\end{sc}
\end{small}
}
\end{center}
% \vskip -0.1in
\caption{Complexity, forward time, and accuracy of different models}
\label{table:cls_time}
\end{table}

\subsection{More Experiments on ModelNet40}
\label{sec:effective_cls}
We also experiment with various settings of our model on the ModelNet40~\cite{wu20153d} dataset. In particular, we analyze the effectiveness of the different distance metrics, explicit usage of $\x_i-\x_j$, and more points.

Table~\ref{table:cls_eff} shows the results. ``Centralization'' denotes using concatenation of $\x_i$ and $\x_i-\x_j$ as the edge features rather than concatenating $\x_i$ and $\x_j$. ``Dynamic graph recomputation'' denotes we reconstruct the graph rather than using a fixed graph. Explicitly centralizing each patch by using the concatenation of $\x_i$ and $\x_i-\x_j$ leads to about 0.5\% improvement for overall accuracy. By dynamically updating graph, there is about 0.7\% improvement, and Figure~\ref{fig:feature_closeness} also suggests that the model can extract semantically meanigful features. Using more points further improves the overall accuracy by 0.6\%.

We also experiment with different numbers $k$ of nearest neighbors as shown in Table~\ref{table:cls_diff_k}. For all experiments, the number of points is still 1024. While we do not exhaustively experiment with all possible $k$, we find with large $k$ that the performance degenerates. This confirms  our hypothesis that for certain density, with large $k$ the Euclidean distance fails to approximate geodesic distance, destroying the geometry of each patch.

\begin{figure}[t!]
  \centering
  \includegraphics[width=0.23\textwidth]{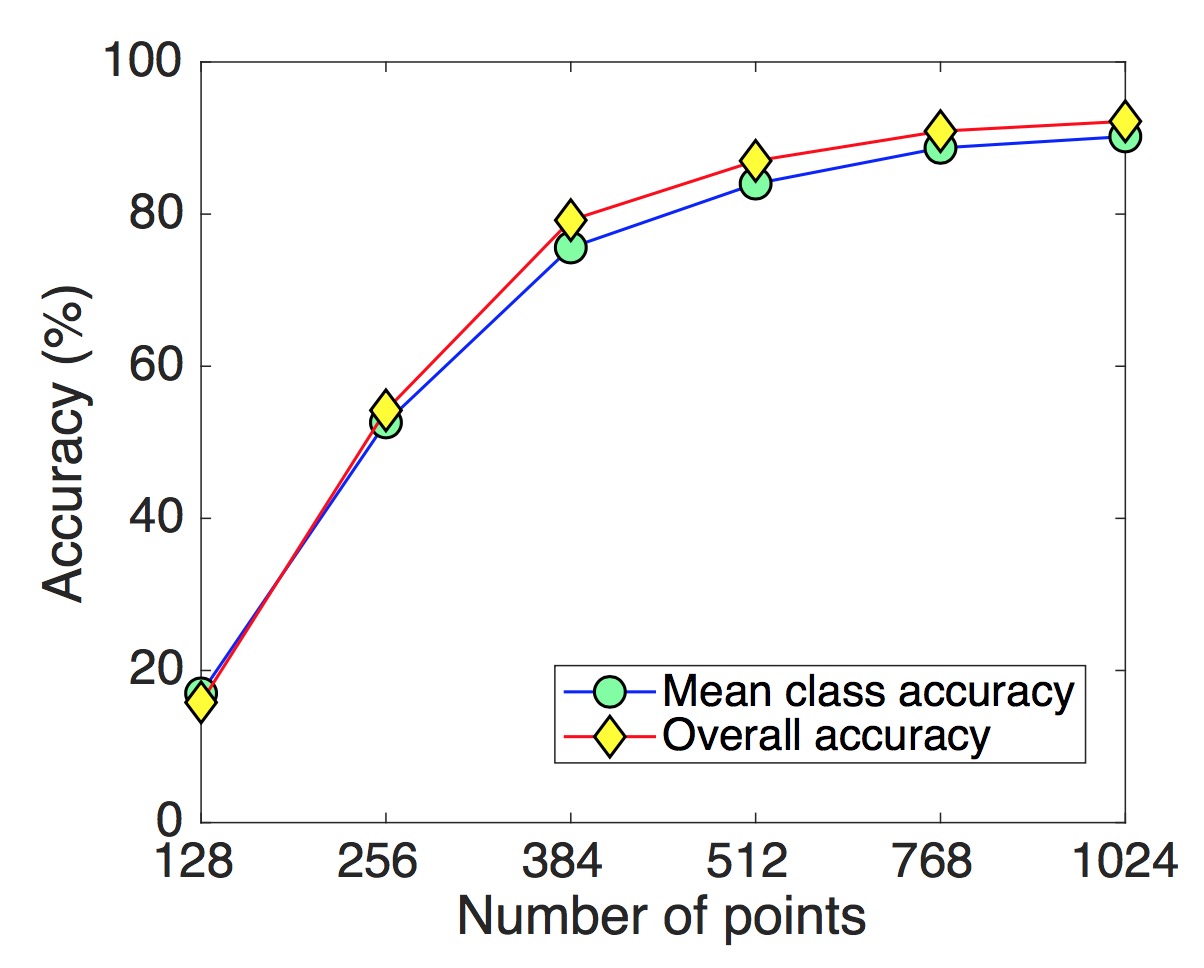}
  \includegraphics[width=0.23\textwidth]{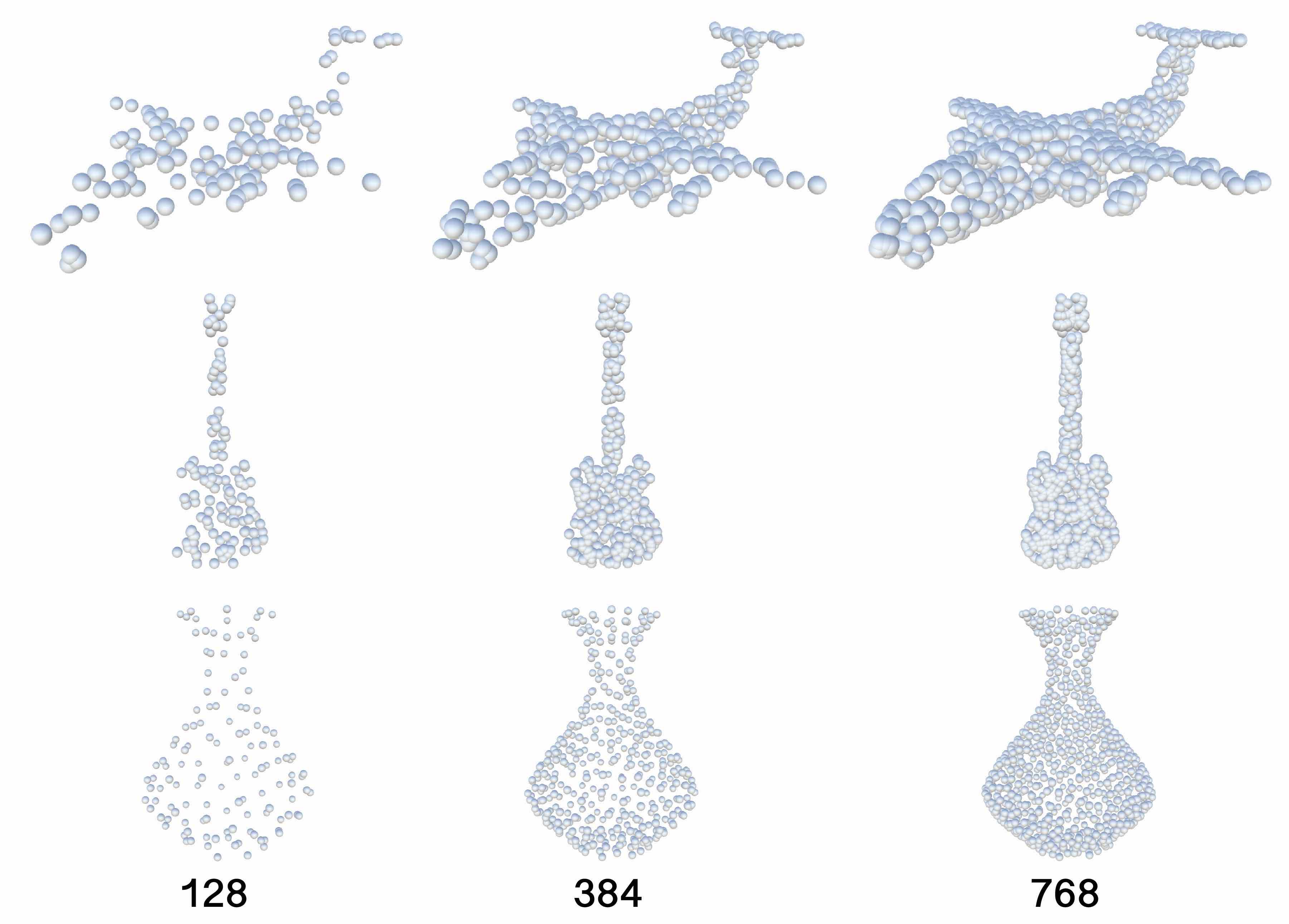}
  \caption{\textbf{Left:} Results of our model tested with random input dropout. The model is trained with number of points being 1024 and $k$ being 20. \textbf{Right:} Point clouds with different number of points. The numbers of points are shown below the bottom row. }
  \label{fig:num_point_acc}
\end{figure}

We further evaluate the robustness of our model (trained on 1,024 points with $k=20$) to point cloud density. We simulate the environment that random input points drops out during testing. Figure~\ref{fig:num_point_acc} shows that even half of points is dropped, the model still achieves reasonable results. With fewer than 512 points, however, performance degenerates dramatically. 

\begin{table}[H]
\vskip 0.1in
\begin{center}
\resizebox{\columnwidth}{!}{
\begin{small}
\begin{sc}
\begin{tabular}{ccccc}
\toprule
 CENT & DYN & MPOINTS & Mean Class Accuracy(\%) & Overall Accuracy(\%) \\
\midrule
 & & & 88.9 & 91.7 \\
x &  &  & 89.3 & 92.2\\
x & x &  & 90.2 & 92.9\\
x & x & x & 90.7 & 93.5\\

\bottomrule
\end{tabular}
\end{sc}
\end{small}
}
\end{center}
\caption{Effectiveness of different components. CENT denotes centralization, DYN denotes dynamical graph recomputation, and MPOINTS denotes experiments with 2048 points}.
\label{table:cls_eff}
\end{table}

\begin{table}[H]
\vskip 0.1in
\begin{center}
\resizebox{\columnwidth}{!}{
\begin{small}
\begin{sc}
\begin{tabular}{lccr}
\toprule
Number of nearest neighbors (k)  & Mean & Overall  \\
   &  Class Accuracy(\%) & Accuracy(\%) \\
\midrule
5 & 88.0 & 90.5 \\
10 & 88.9 & 91.4 \\
20 & 90.2 & 92.9 \\
40 & 89.4 & 92.4\\
\bottomrule
\end{tabular}
\end{sc}
\end{small}
}
\end{center}
% \vskip -0.1in
\caption{Results of our model with different numbers of nearest neighbors.}
\label{table:cls_diff_k}
\end{table}

% many part segs
\begin{figure}[t!]
  \centering
  \includegraphics[width=0.5\textwidth]{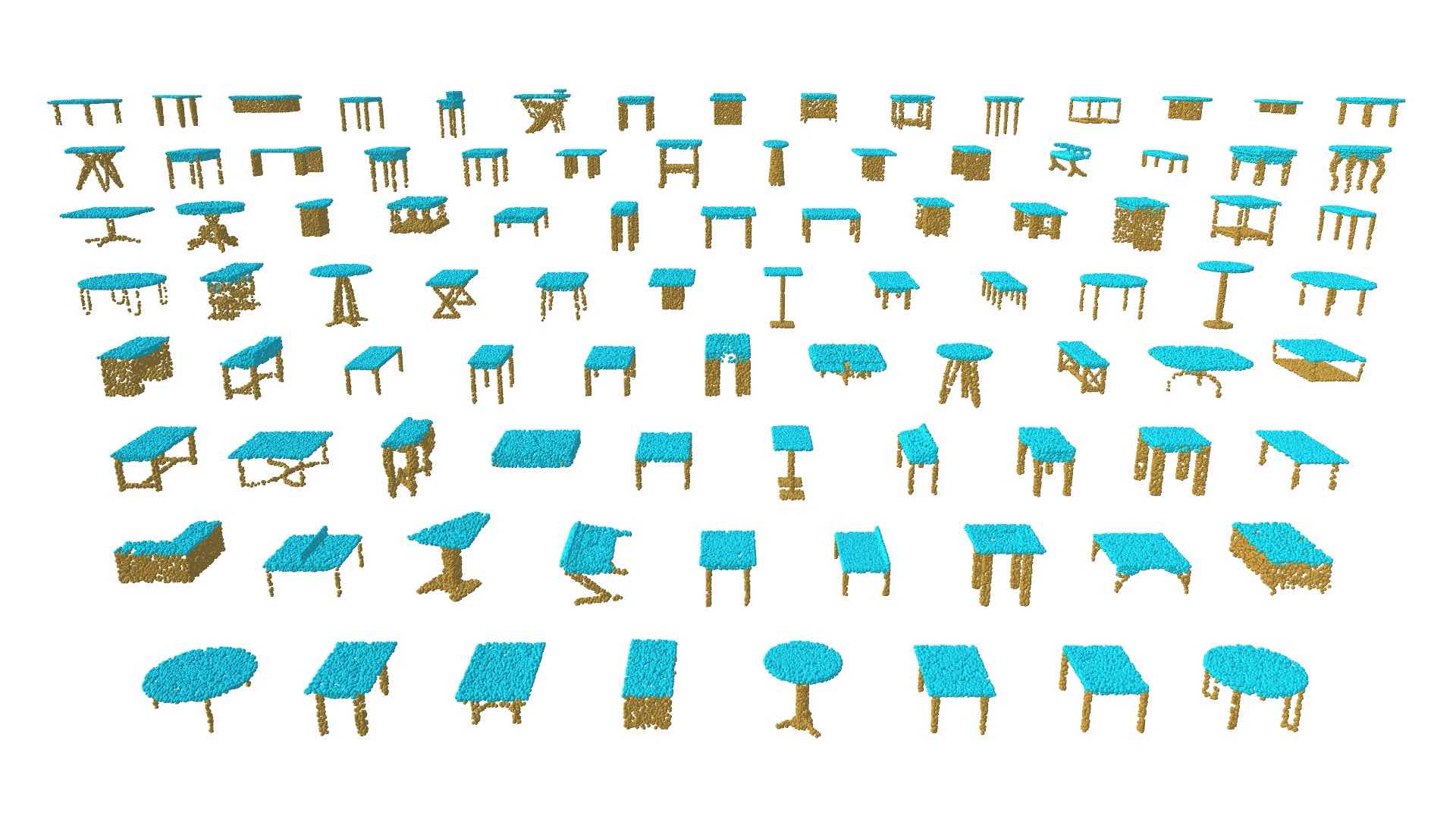}\vspace{-.3in}
  \includegraphics[width=0.5\textwidth]{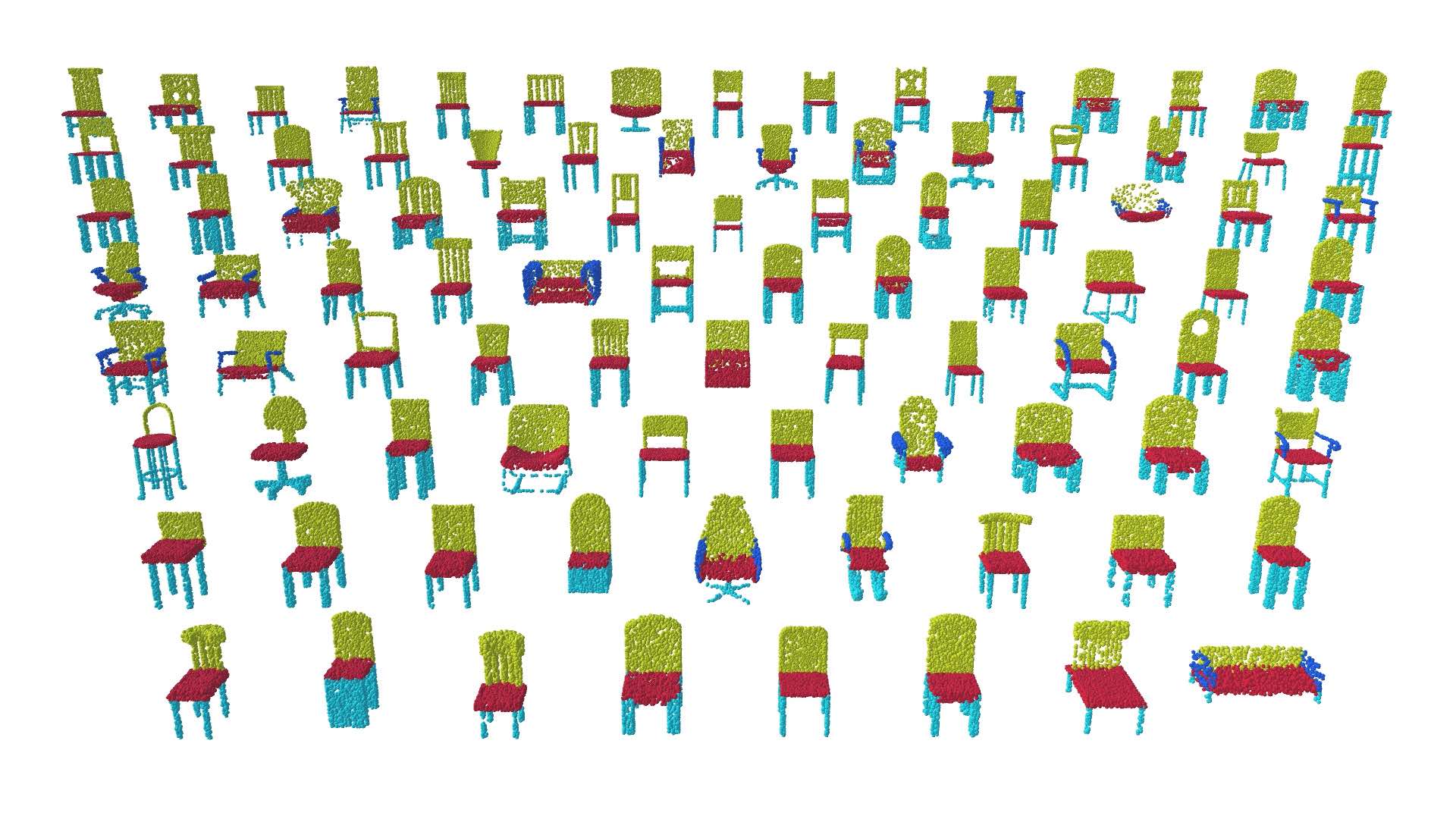}\vspace{-.3in}
  \includegraphics[width=0.5\textwidth]{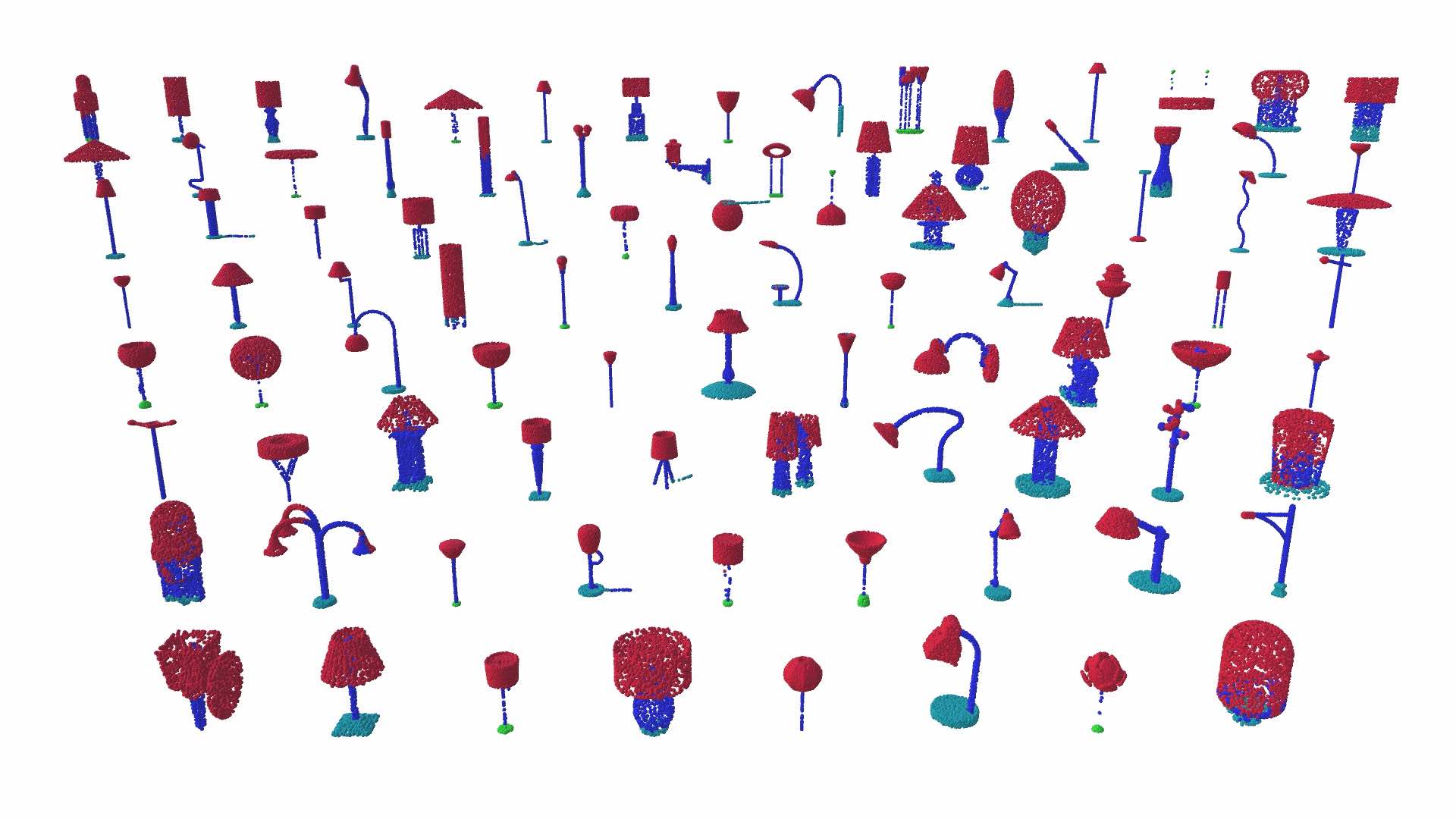}\vspace{-.3in}
  \caption{Our part segmentation testing results for tables, chairs and lamps.}
  \label{fig:part_more}
\end{figure}

\subsection{Part Segmentation}
\label{sec:part_seg}
\paragraph{Data} We extend our EdgeConv model architectures for part segmentation task on ShapeNet part dataset \cite{yi2016scalable}.
For this task, each point from a point cloud set is classified into one of a few predefined part category labels.
The dataset contains 16,881 3D shapes from 16 object categories, annotated with 50 parts in total.
2,048 points are sampled from each training shape, and most sampled point sets are labeled with less than six parts. 
We follow the official train$/$validation$/$test split scheme as \citet{chang2015shapenet} in our experiment.

\paragraph{Architecture}
The network architecture is illustrated in Figure~\ref{fig:model_architecture} (bottom branch). After a spatial transformer network, three EdgeConv layers are used. A shared fully-connected layer $(1024)$ aggregates information from the previous layers. Shortcut connections are used to include all the EdgeConv outputs as local feature descriptors. At last, three shared fully-connected layers $(256, 256, 128)$ are used to transform the pointwise features. Batch-norm, dropout, and ReLU are included in the similar fashion to our classification network.

\paragraph{Training} The same training setting as in our classification task is adopted. A distributed training scheme is further implemented on two NVIDIA TITAN X GPUs to maintain the training batch size.

\begin{figure}[t!]
  \centering
  \includegraphics[width=0.15\textwidth]{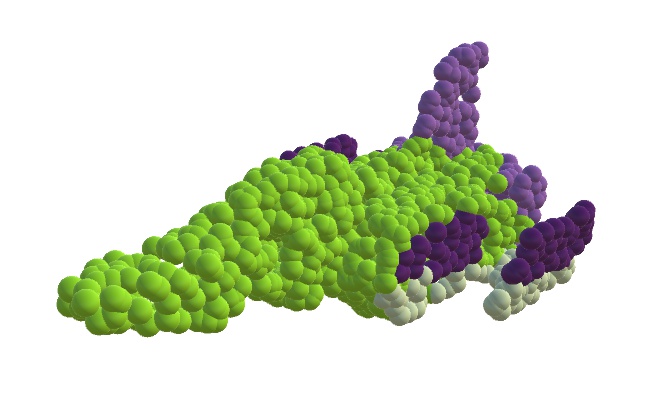} 
  \includegraphics[width=0.15\textwidth]{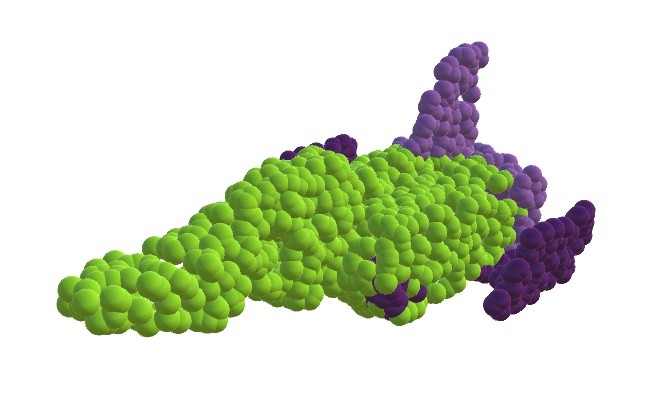} 
  \includegraphics[width=0.15\textwidth]{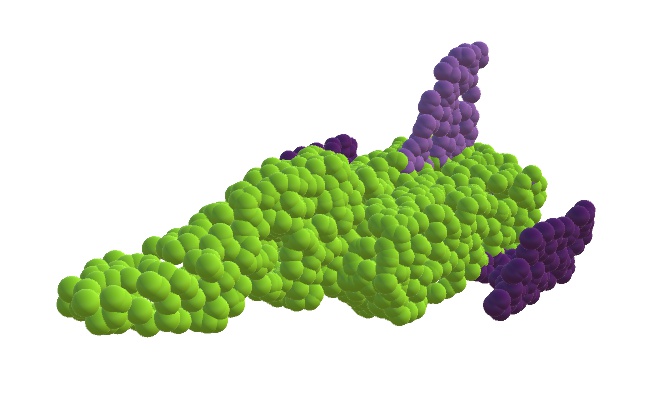}
  \includegraphics[width=0.15\textwidth]{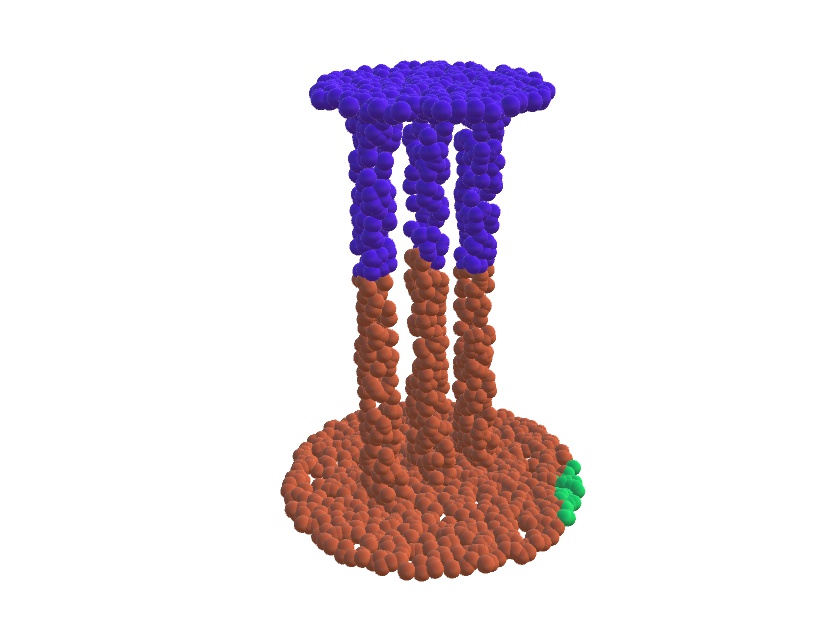} 
  \includegraphics[width=0.15\textwidth]{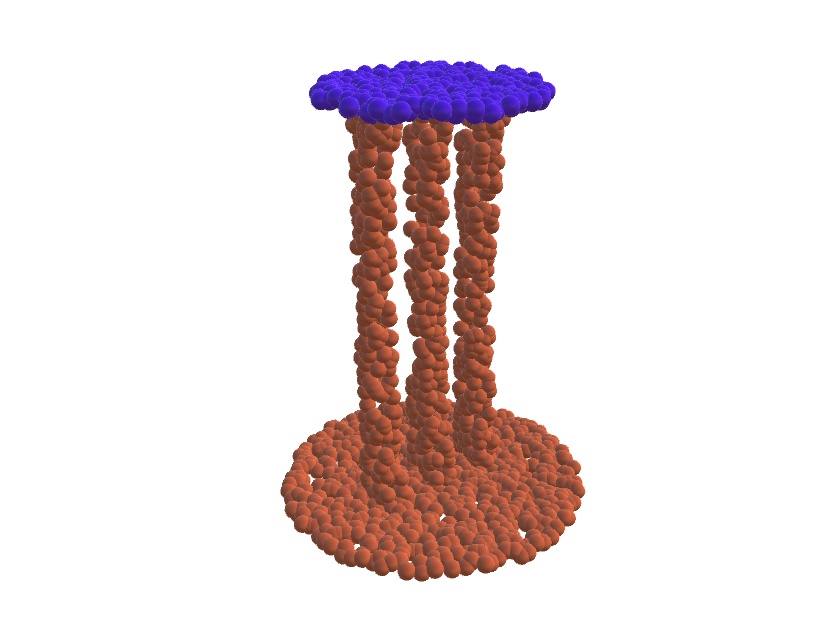} 
  \includegraphics[width=0.15\textwidth]{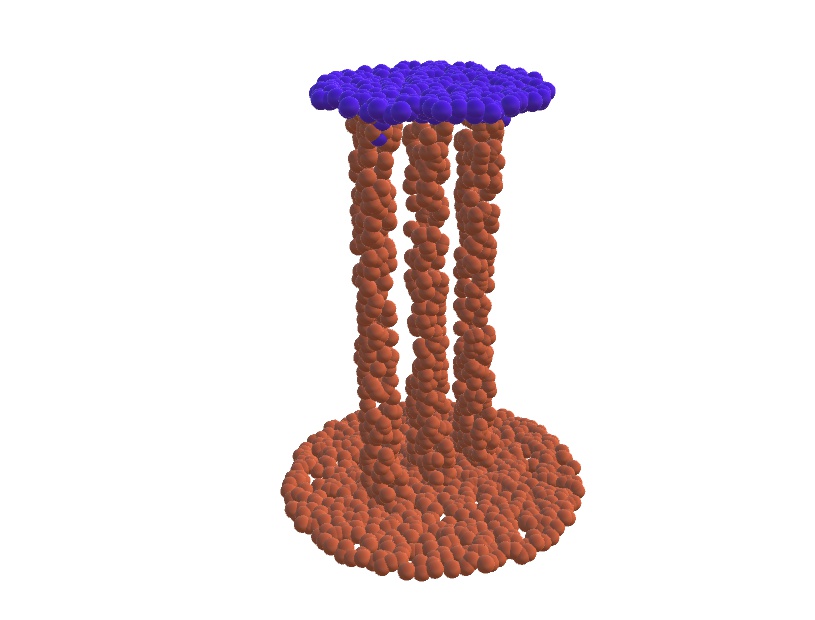} 
  \includegraphics[width=0.15\textwidth]{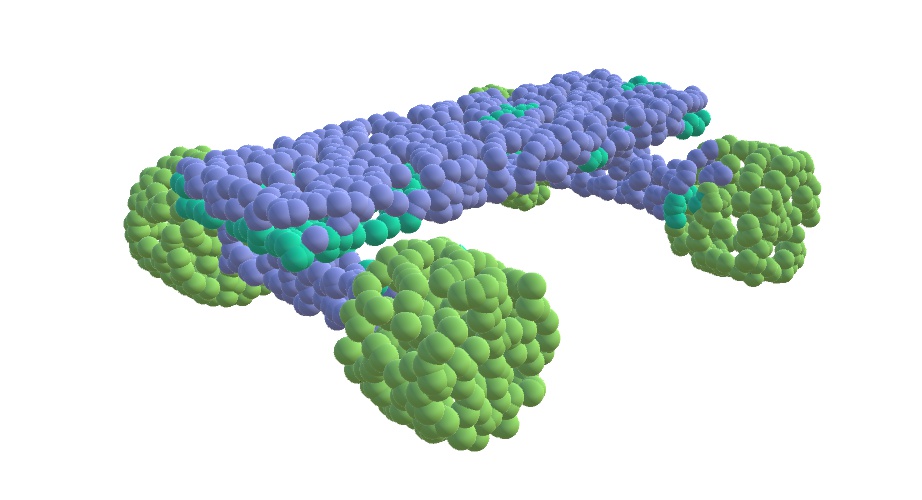}
  \includegraphics[width=0.15\textwidth]{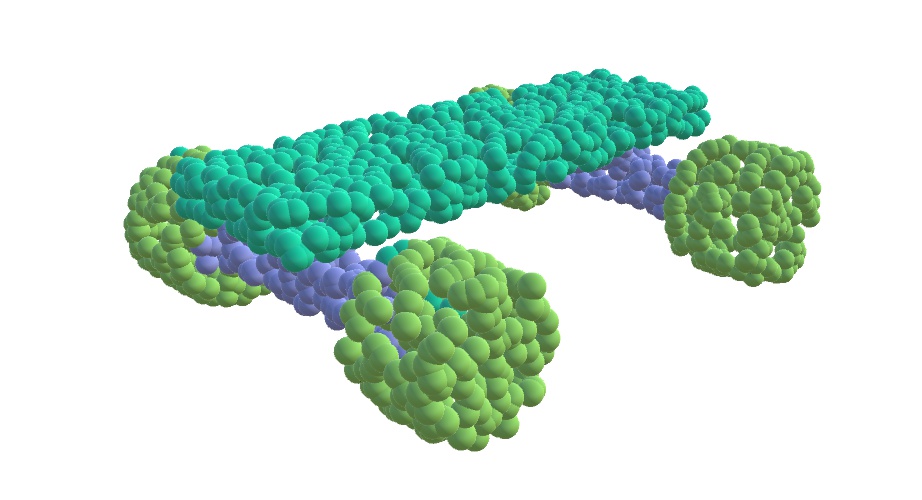} 
  \includegraphics[width=0.15\textwidth]{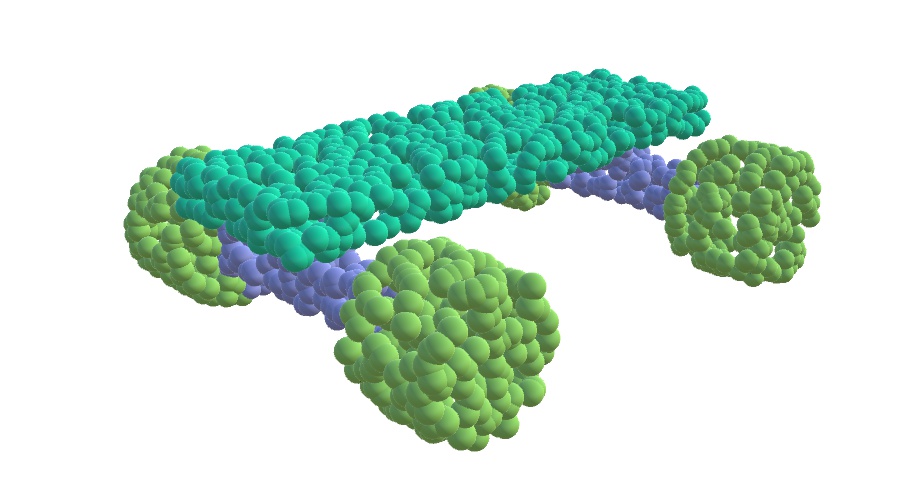} 
  \includegraphics[width=0.15\textwidth]{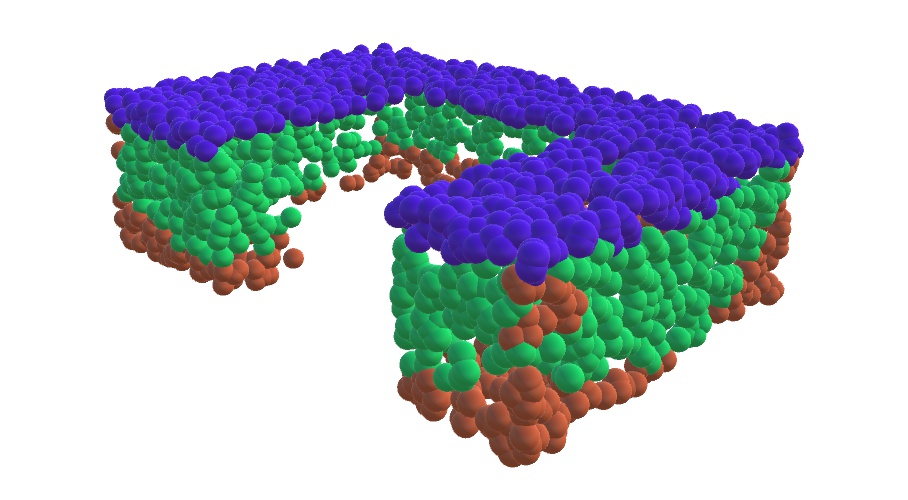} 
  \includegraphics[width=0.15\textwidth]{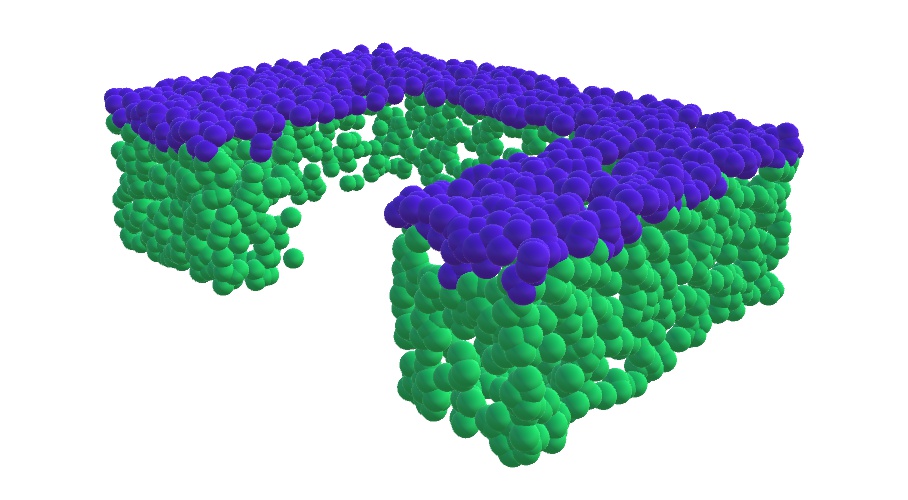} 
  \includegraphics[width=0.15\textwidth]{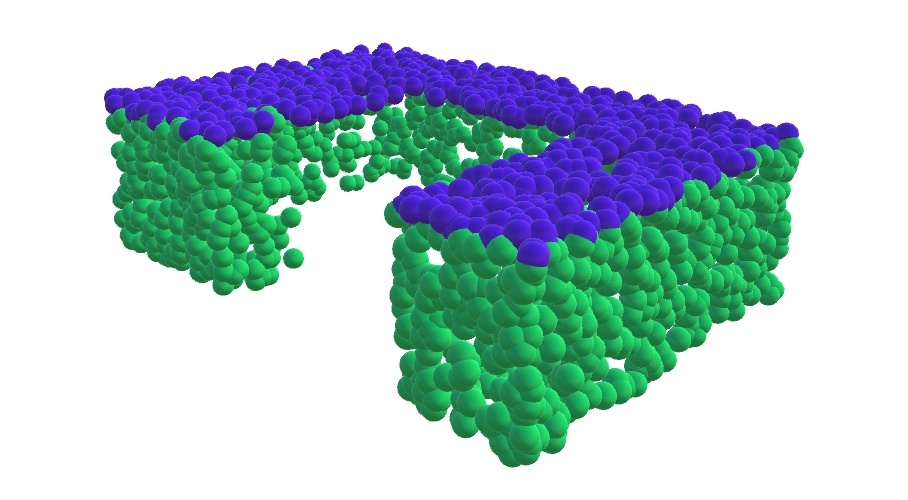} 
  \includegraphics[width=0.15\textwidth]{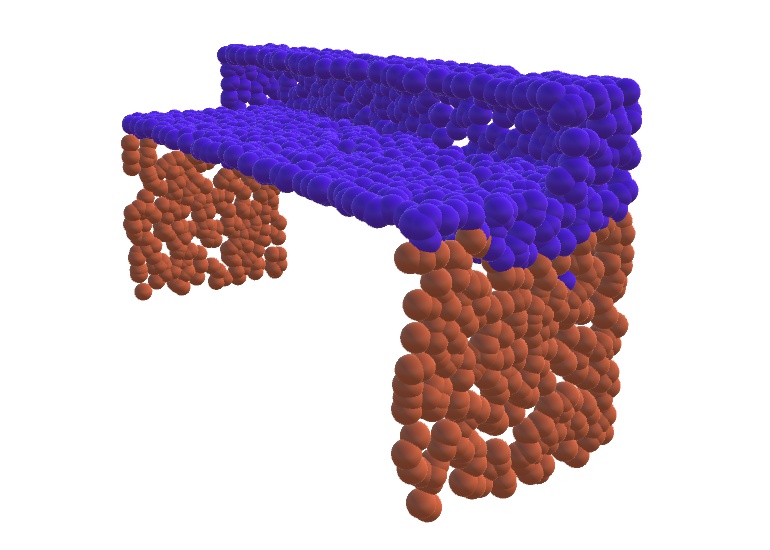} 
  \includegraphics[width=0.15\textwidth]{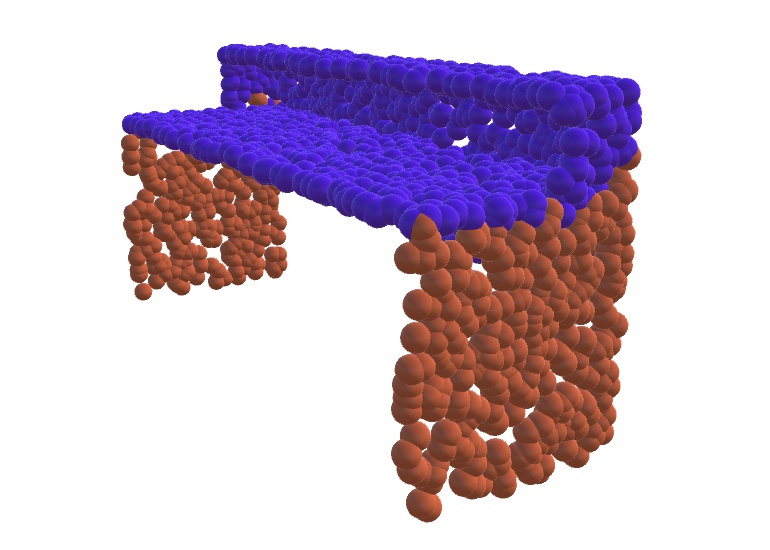} 
  \includegraphics[width=0.15\textwidth]{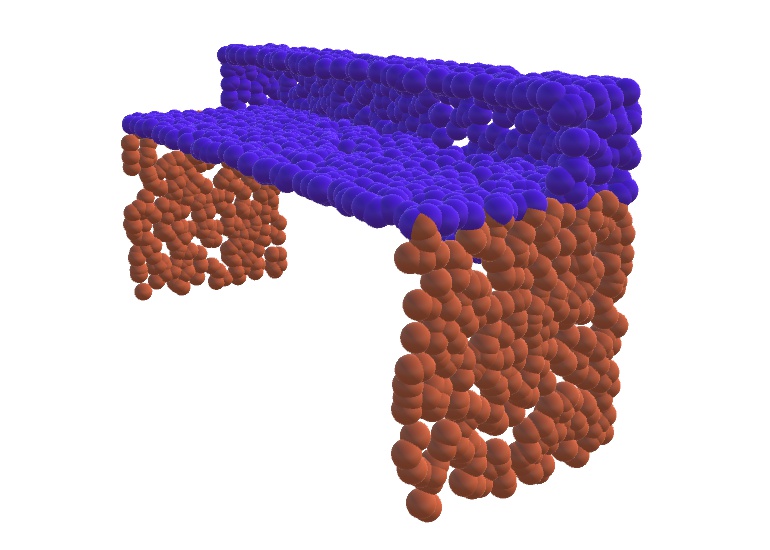}
 
  \hspace{0.01\textwidth} {PointNet} \hspace{0.1\textwidth}  {Ours} \hspace{0.08\textwidth}  {Ground truth}
  
  \caption{Compare part segmentation results. For each set, from left to right: PointNet, ours and ground truth.}
  \label{fig:part_segmentation}
\end{figure}

\paragraph{Results} We use Intersection-over-Union (IoU) on points to evaluate our model and compare with other benchmarks.
We follow the same evaluation scheme as PointNet: The IoU of a shape is computed by averaging the IoUs of different parts occurring in that shape, and
the IoU of a category is obtained by averaging the IoUs of all the shapes belonging to that category.
The mean IoU (mIoU) is finally calculated by averaging the IoUs of all the testing shapes.
We compare our results with PointNet \cite{DBLP:journals/corr/QiSMG16}, PointNet++ \cite{DBLP:journals/corr/QiYSG17}, Kd-Net \cite{klokov2017escape}, LocalFeatureNet \cite{shen2017neighbors}, PCNN \cite{pcnn2018}, and PointCNN \cite{pointcnn}.
The evaluation results are shown in Table \ref{table:part_segmentation}.
We also visually compare the results of our model and PointNet in Figure \ref{fig:part_segmentation}. More examples are shown in Figure~\ref{fig:part_more}.

\begin{figure}[h!]
  \centering
  \includegraphics[width=0.5\textwidth]{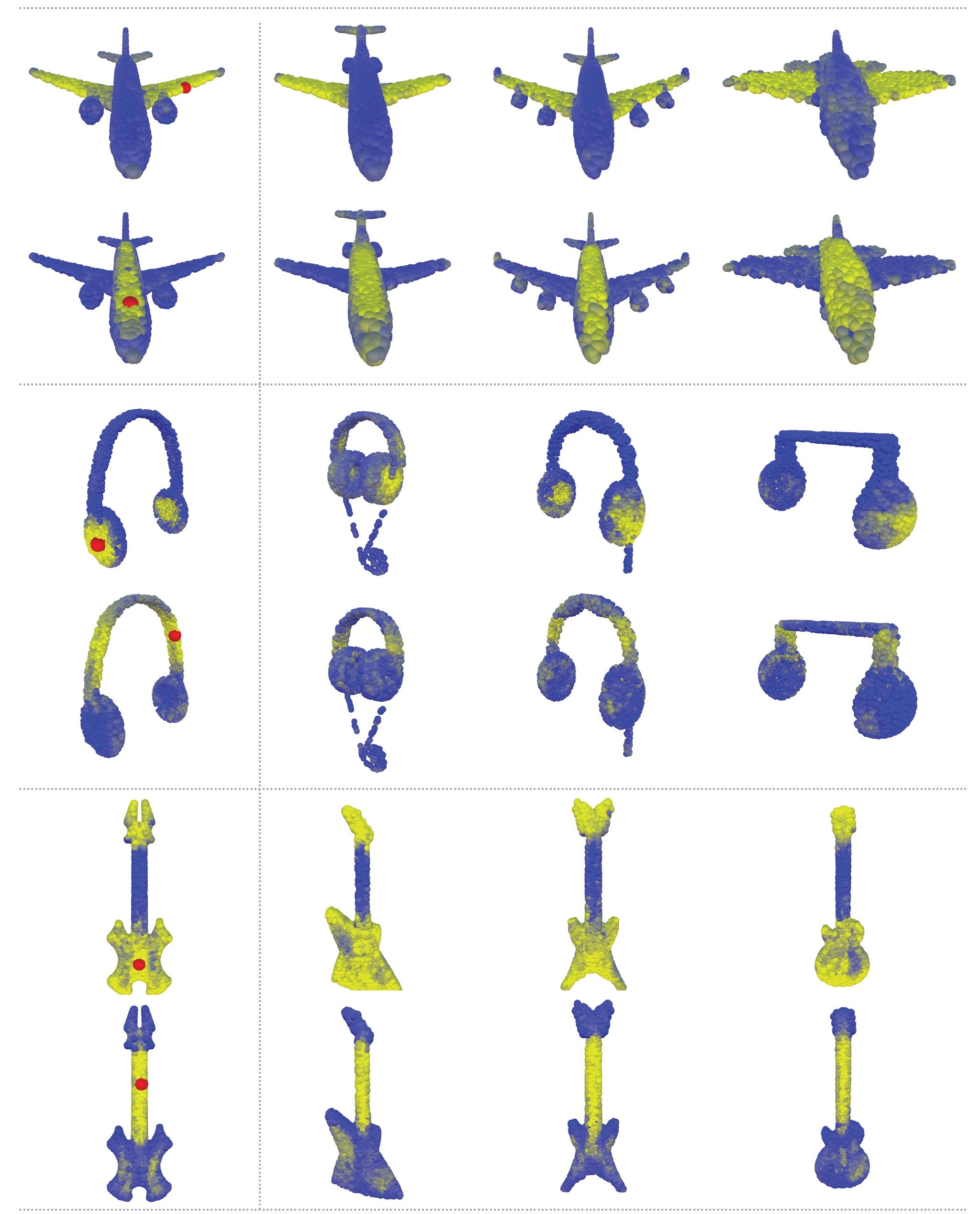} 
  \hspace{0.05\textwidth} {Source points} \hspace{0.03\textwidth}  {Other point clouds from the same category } 
  
  \caption{Visualize the Euclidean distance (yellow: near, blue: far) between source points (red points in the left column) and multiple point clouds from the same category in the feature space after the third EdgeConv layer. Notice source points not only capture semantically similar structures in the point clouds that they belong to, but also capture semantically similar structures in other point clouds from the same category.}
  \label{fig:cross_part}
\end{figure}

\paragraph{Intra-cloud distances} We next explore the relationships between different point clouds captured using our features. As shown in Figure~\ref{fig:cross_part}, we take one red point from a source point cloud and compute its distance in feature space to points in other point clouds from the same category. An interesting finding is that although points are from different sources, they are close to each other if they are from semantically similar parts. We evaluate on the features after the third layer of our segmentation model for this experiment. 

\paragraph{Segmentation on partial data} 

\begin{figure}[t!]
  \centering
  \includegraphics[width=0.2\textwidth]{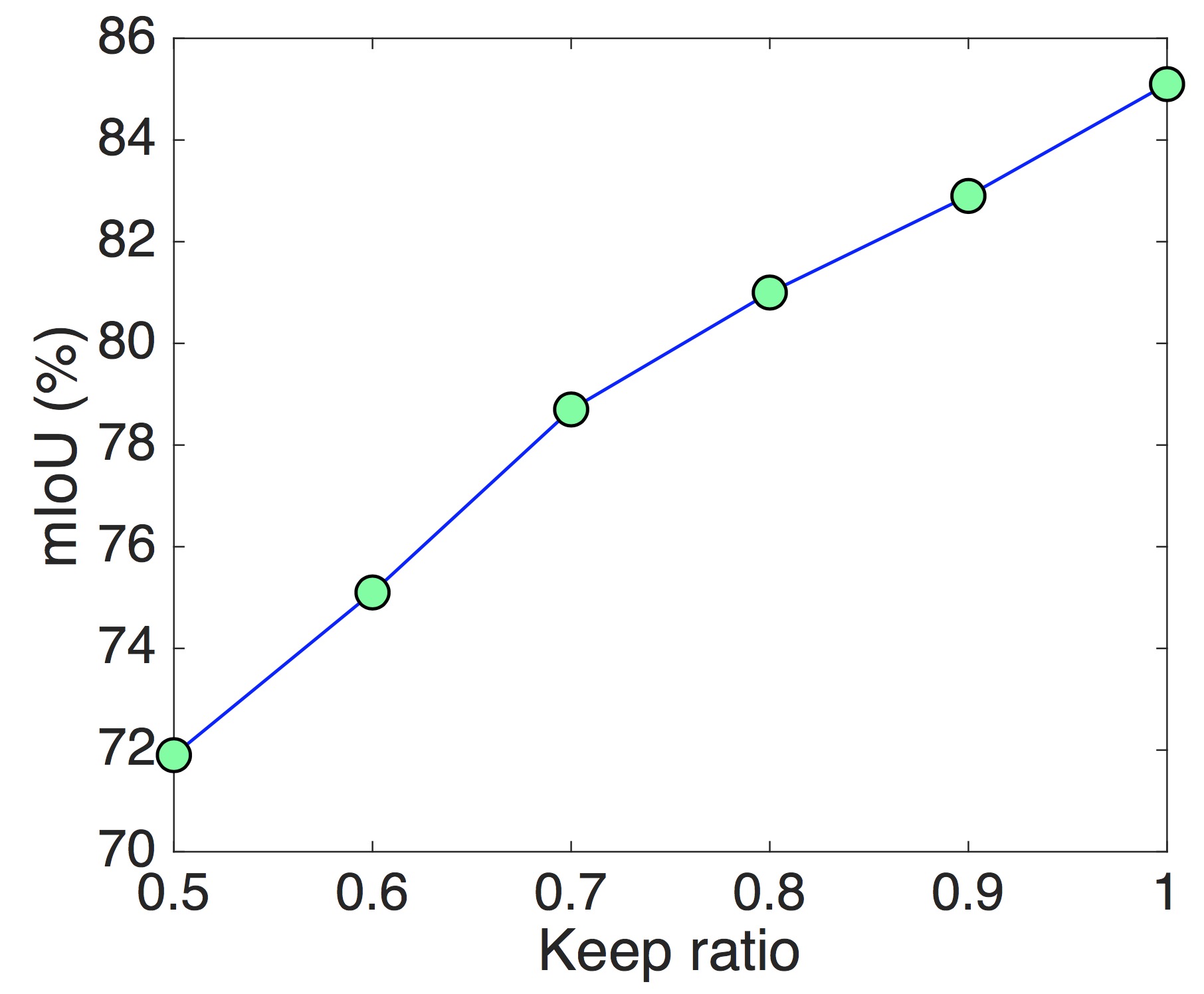} 
  \includegraphics[width=0.26\textwidth]{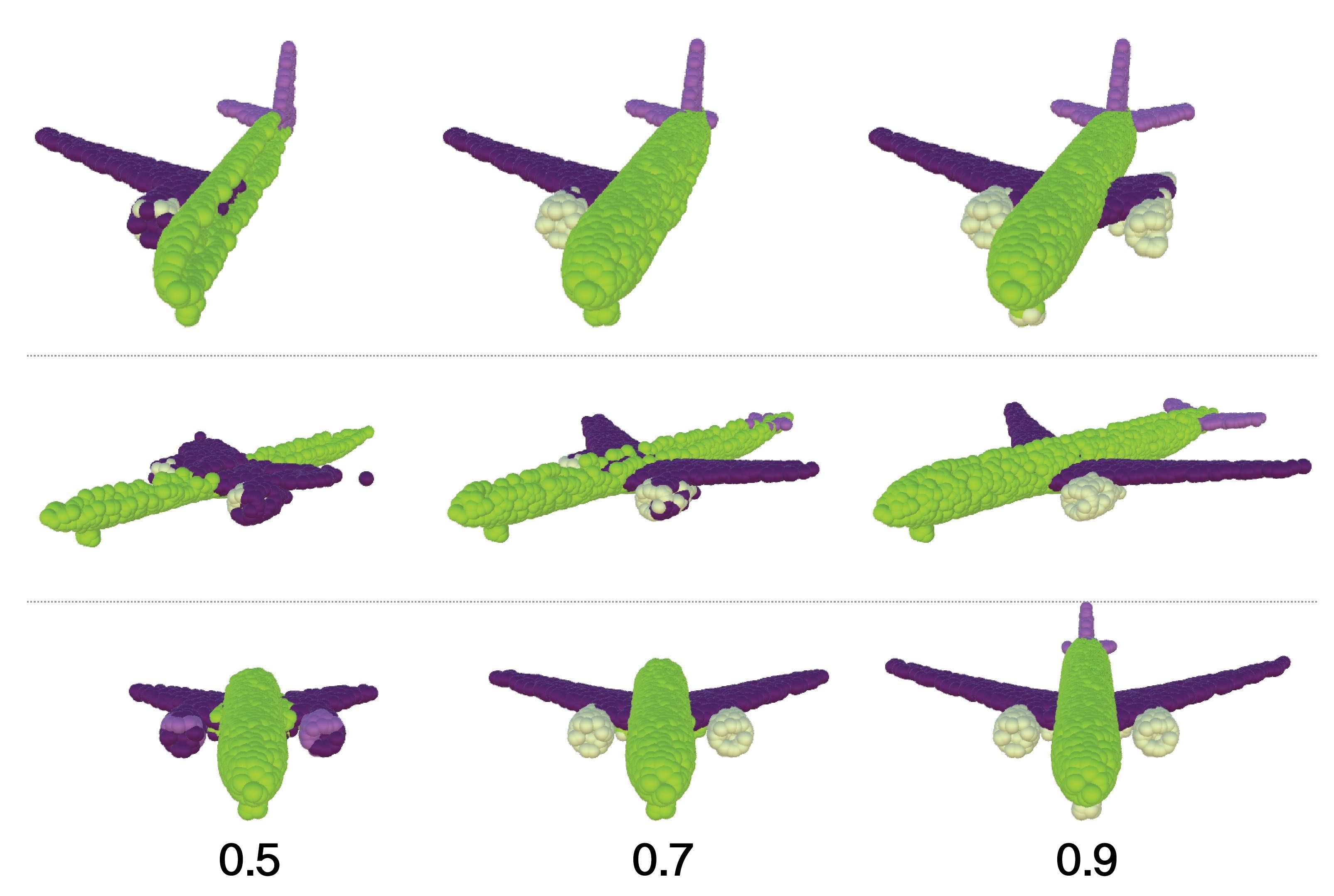} 
  \caption{\textbf{Left:} The mean IoU (\%) improves when the ratio of kept points increases. Points are dropped from one of six sides (top, bottom, left, right, front and back) randomly during evaluation process. \textbf{Right:} Part segmentation results on partial data. Points on each row are dropped from the same side. The keep ratio is shown below the bottom row. Note that the segmentation results of turbines are improved when more points are included.}
  \label{fig:partial_part}
\end{figure}

Our model is robust to partial data. We simulate the environment that part of the shape is dropped from one of six sides (top, bottom, right, left, front and back) with different percentages. The results are shown in Figure~\ref{fig:partial_part}. On the left, the mean IoU versus ``keep ratio'' is shown. On the right, the results for an airplane model are visualized.

% part seg comp.

% cross part.

% partial part

\begin{table*}[t!]
\vskip 0.1in
\begin{center}
\resizebox{2.1\columnwidth}{!}{
\begin{small}
\begin{sc}
\begin{tabular}{l|c|ccccccccccccccccr}
\toprule
                & \textbf{mean} & areo & bag & cap & car & chair & ear & guitar & knife & lamp & laptop & motor & mug & pistol & rocket & skate & table  \\
                &          &          &.       &        &       &          & phone&       &          &          &           &            &         &           &           & board &          \\
\midrule
 \# shapes   &       &2690 &76  &55 & 898 & 3758 & 69 & 787 & 392 & 1547 & 451 & 202 & 184 & 283 & 66 & 152 & 5271 & \\
\midrule
PointNet              & 83.7 & 83.4 & 78.7 & 82.5 & 74.9 & 89.6 & 73.0 & 91.5 & 85.9 & 80.8 & 95.3 & 65.2 & 93.0 & 81.2 & 57.9 & 72.8 & 80.6 \\
PointNet++          & 85.1 & 82.4 & 79.0 & \textbf{87.7} & 77.3 & \textbf{90.8} & 71.8 & 91.0 & 85.9 & 83.7 & 95.3 & 71.6 & 94.1 & 81.3 & 58.7 & 76.4 & 82.6 \\
Kd-Net                 & 82.3 &  80.1 & 74.6 & 74.3 & 70.3 & 88.6 & 73.5 & 90.2 & 87.2 & 81.0 & 94.9 & 57.4 & 86.7 & 78.1 & 51.8 & 69.9 & 80.3 \\
LocalFeatureNet  & 84.3 & \textbf{86.1} & 73.0 & 54.9 & 77.4 & 88.8 & 55.0 & 90.6 & 86.5 & 75.2 & \textbf{96.1} & 57.3 & 91.7 & 83.1 & 53.9 & 72.5 & \textbf{83.8} \\
PCNN & 85.1 & 82.4 & 80.1 & 85.5 & 79.5 & \textbf{90.8} & 73.2 & 91.3 & 86.0 & 85.0 & 95.7 & 73.2 & 94.8 & 83.3 & 51.0 & 75.0 & 81.8  \\

PointCNN & \textbf{86.1} & 84.1 & \textbf{86.45} & 86.0 & \textbf{80.8} & 90.6 & \textbf{79.7} & \textbf{92.3} & \textbf{88.4} & \textbf{85.3} & \textbf{96.1} & \textbf{77.2} & \textbf{95.3} & \textbf{84.2} & \textbf{64.2} & \textbf{80.0} & 83.0 \\
\midrule
Ours                     & 85.2 & 84.0 & 83.4 & 86.7 & 77.8 & 90.6 & 74.7 & 91.2 & 87.5 & 82.8 & 95.7 & 66.3 & 94.9 & 81.1 & 63.5 & 74.5 & 82.6 \\

\bottomrule
\end{tabular}
\end{sc}
\end{small}
}
\end{center}
% \vskip -0.1in
\caption{Part segmentation results on ShapeNet part dataset. Metric is mIoU(\%) on points.}
\label{table:part_segmentation}
\end{table*}

%%%%%%%%%%%%%%%%%
% % surface normal
% surface normals
% \begin{figure}[h]
%   \centering
%   \includegraphics[width=0.5\textwidth]{./sections/figure/normal_est} 

%   \caption{Surface normal estimation results. The colors shown in the figure are RGB-coded surface normals, meaning XYZ components of surface normal vectors are put into RGB color channels. For each pair: our prediction (left) and ground truth (right).}
%   \label{fig:normal_est}
% \end{figure}

%%%%%%%%%%%%%%%%%

% sem seg
\begin{figure*}[t!]

  \centering

  \includegraphics[width=0.22\textwidth]{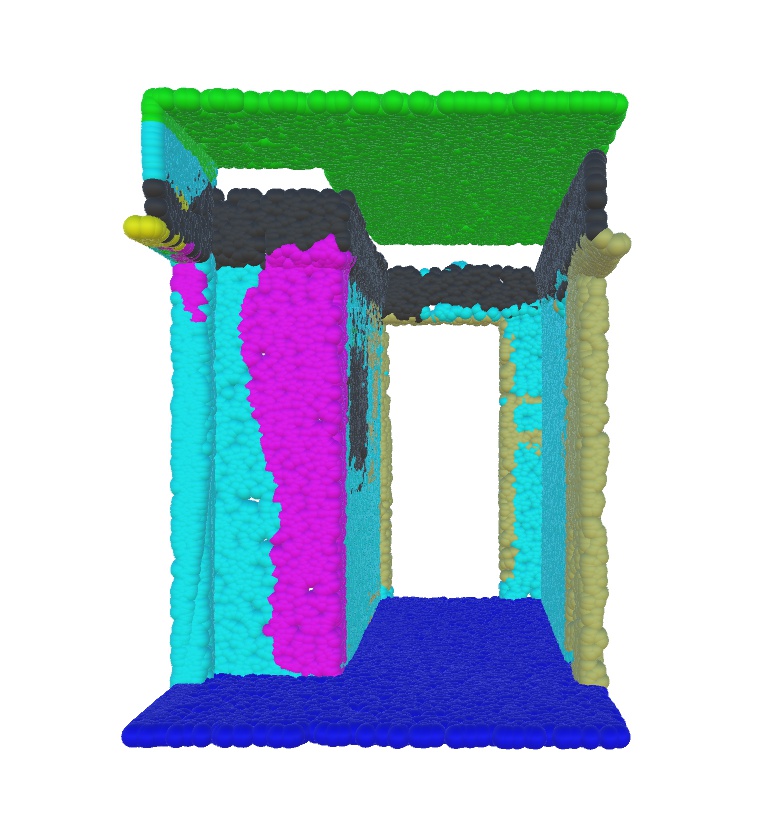}
  \includegraphics[width=0.22\textwidth]{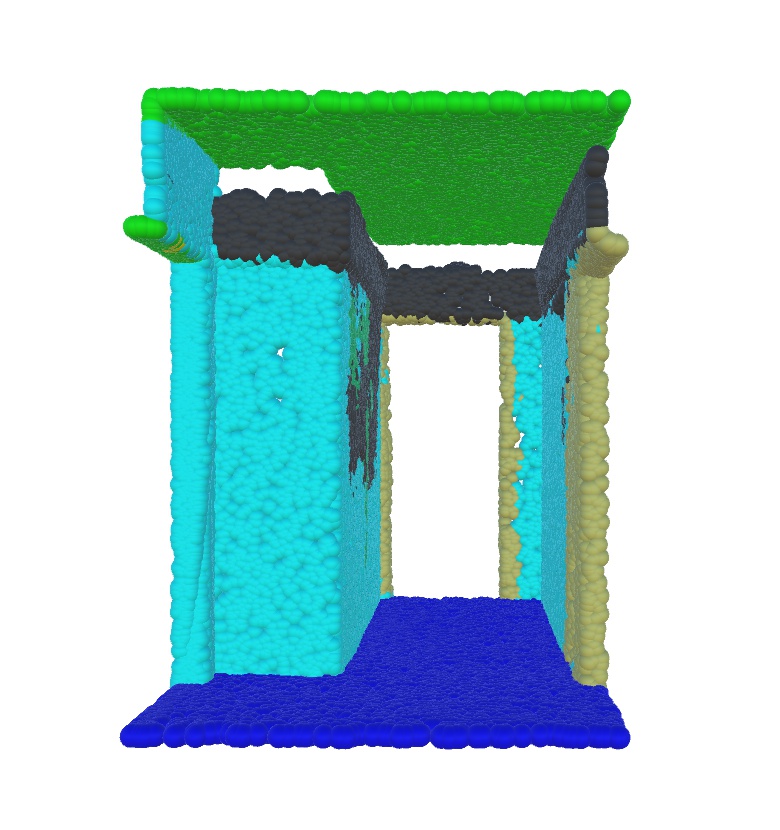}
  \includegraphics[width=0.22\textwidth]{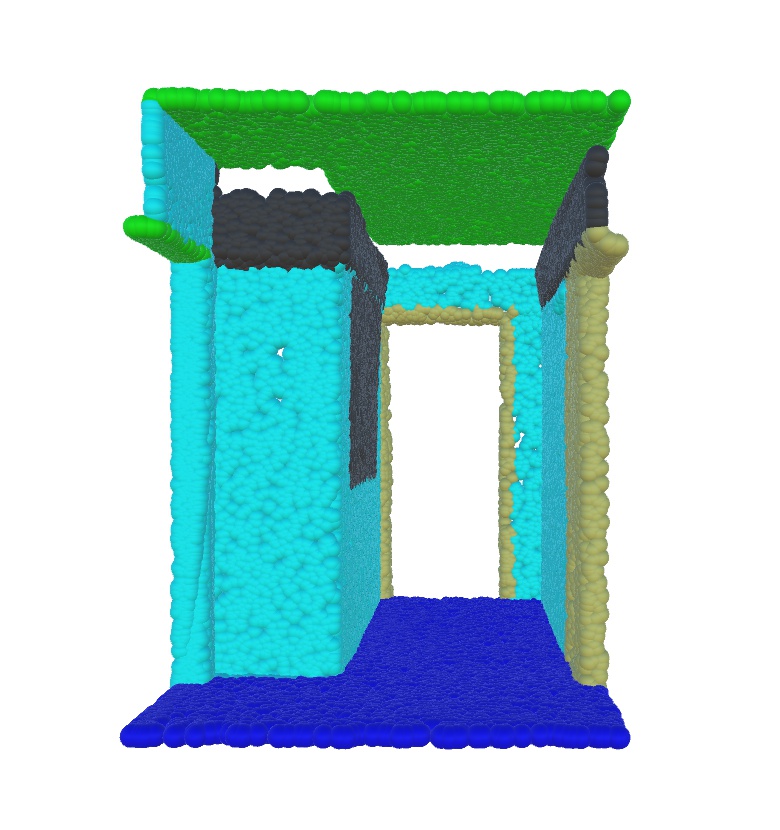}
  \includegraphics[width=0.22\textwidth]{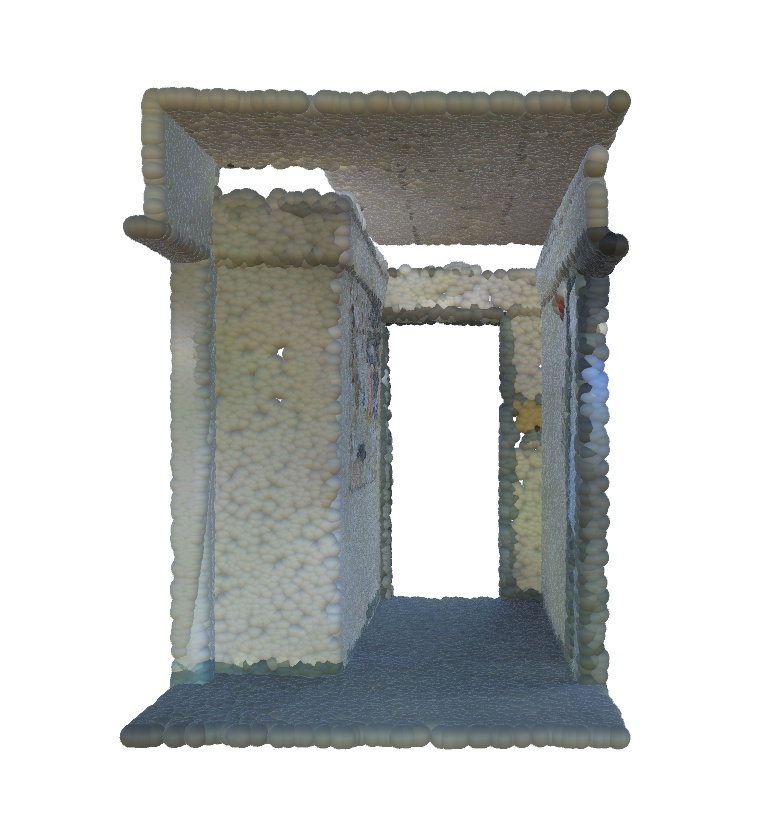}
  
  \includegraphics[width=0.22\textwidth]{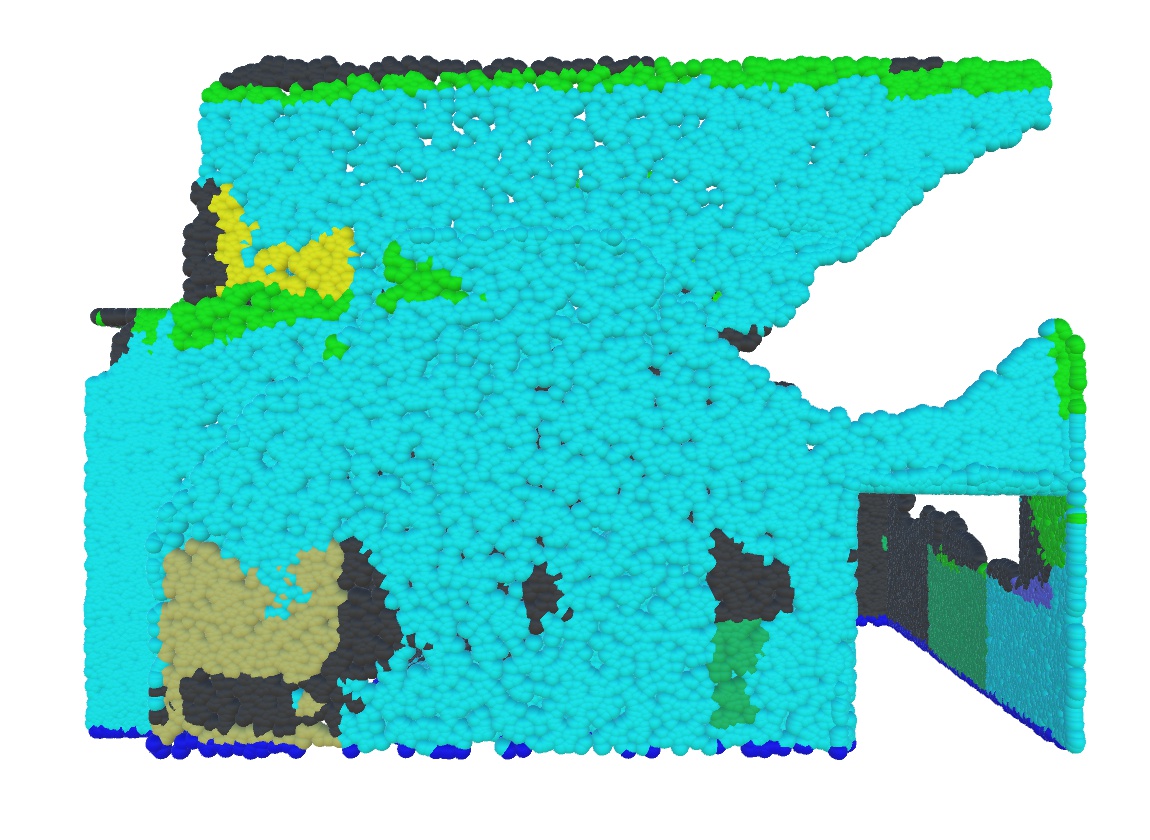}
  \includegraphics[width=0.22\textwidth]{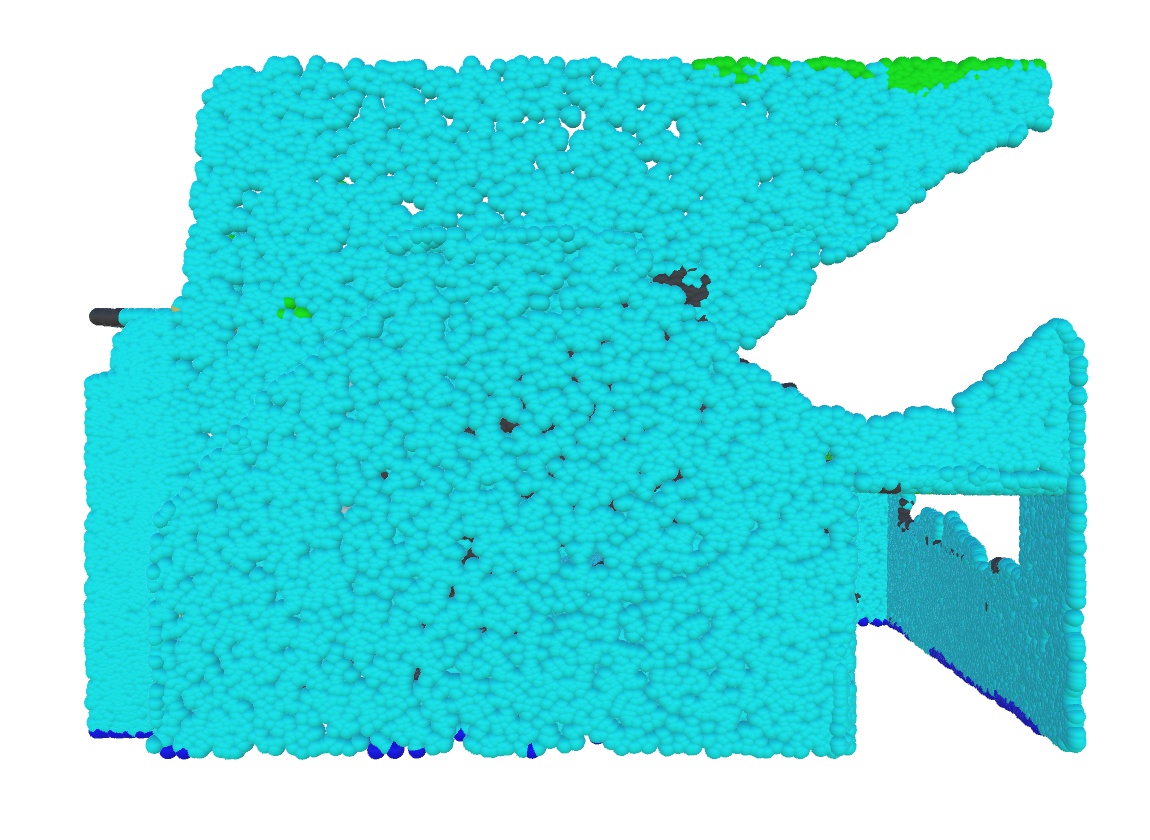}
  \includegraphics[width=0.22\textwidth]{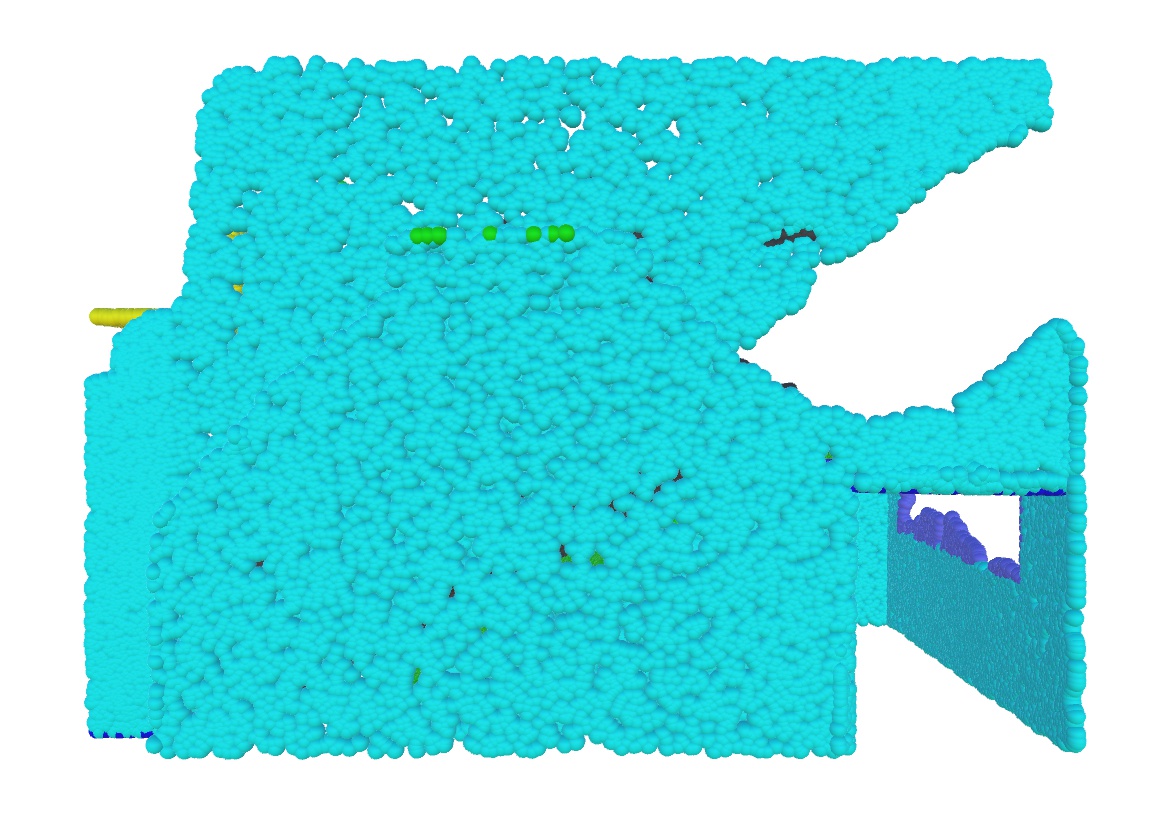}
  \includegraphics[width=0.22\textwidth]{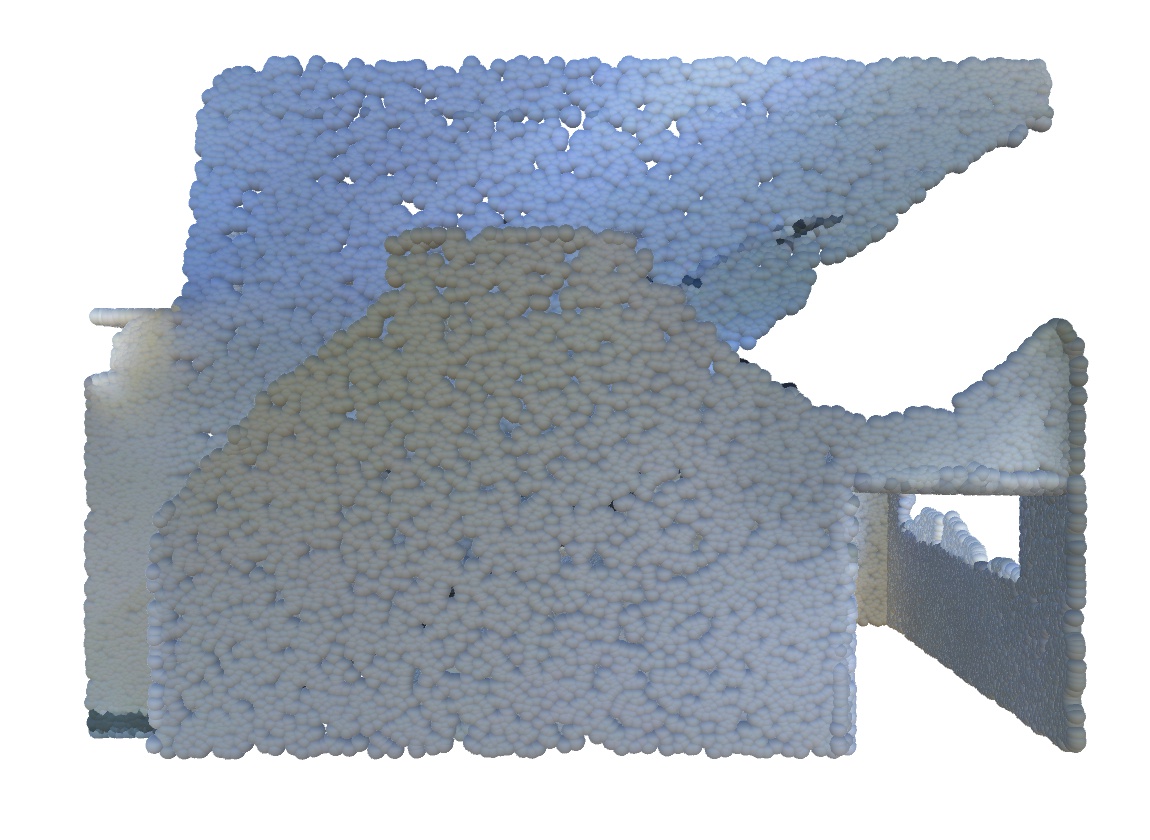}
  
  \includegraphics[width=0.22\textwidth]{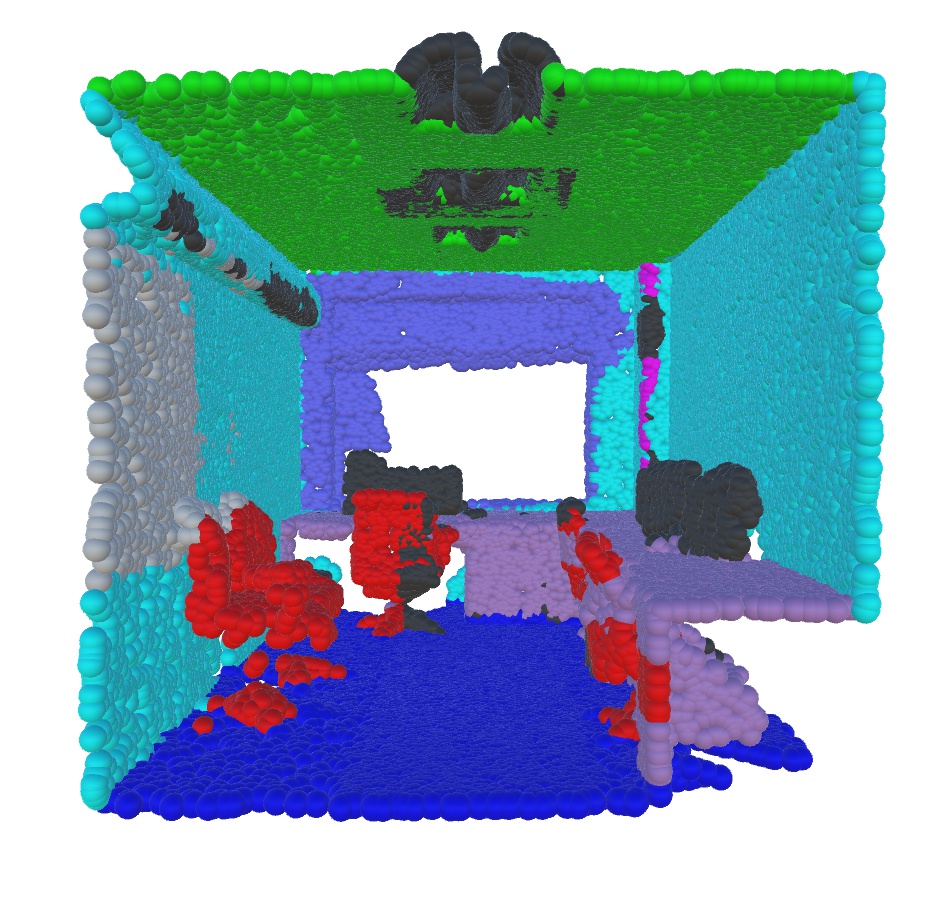}
  \includegraphics[width=0.22\textwidth]{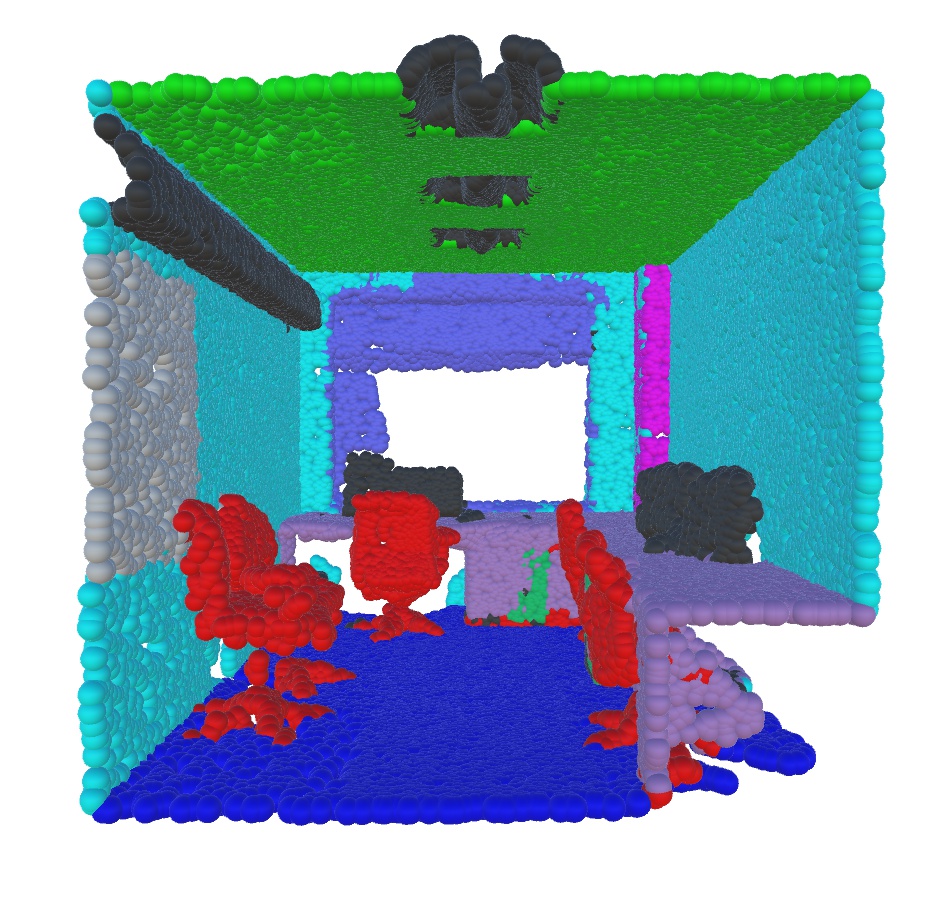}
  \includegraphics[width=0.22\textwidth]{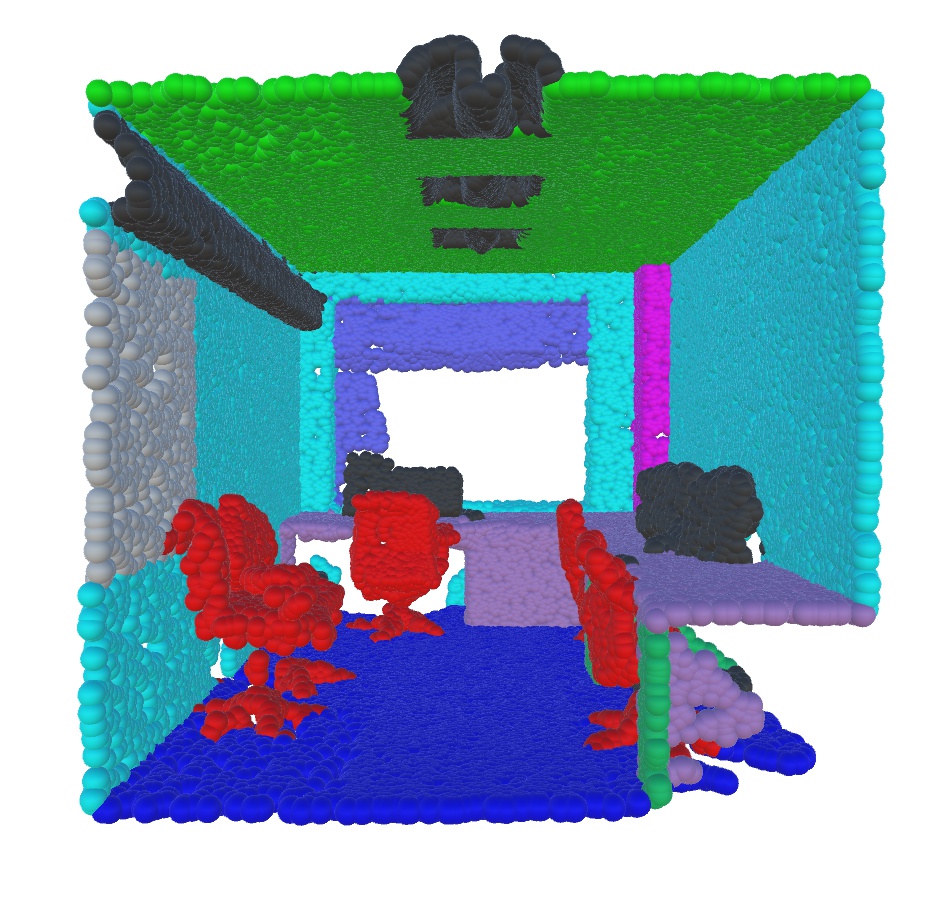}
  \includegraphics[width=0.22\textwidth]{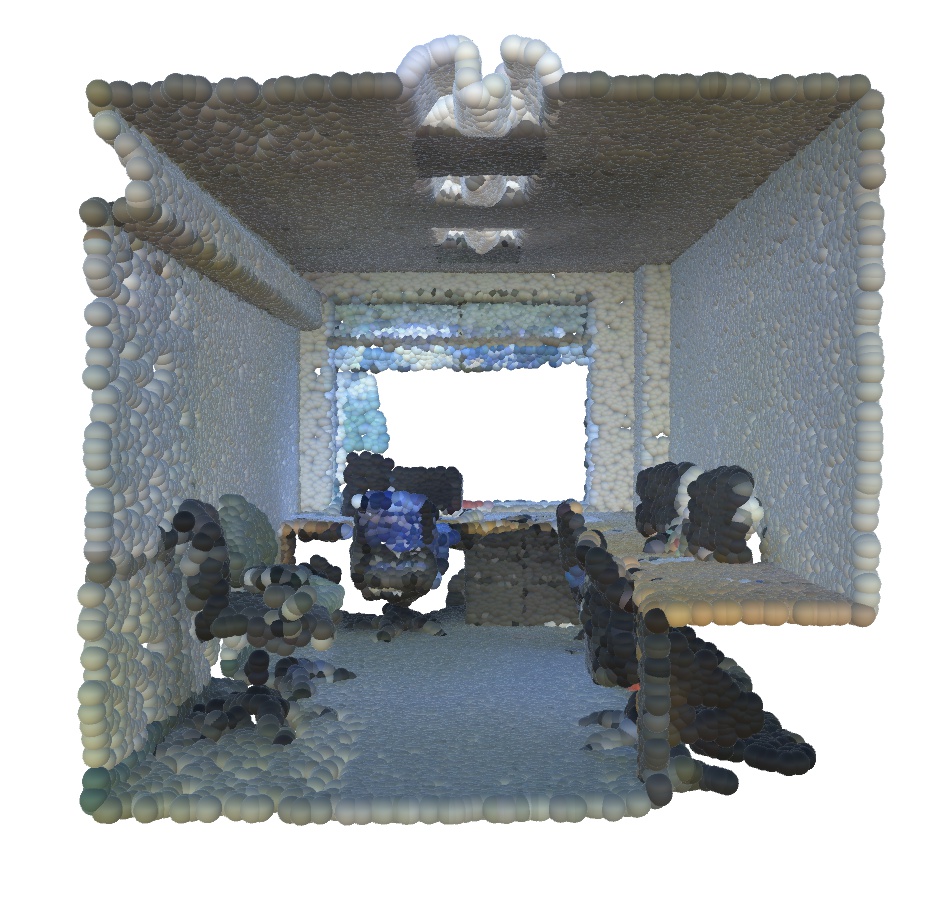}
  
  \includegraphics[width=0.22\textwidth]{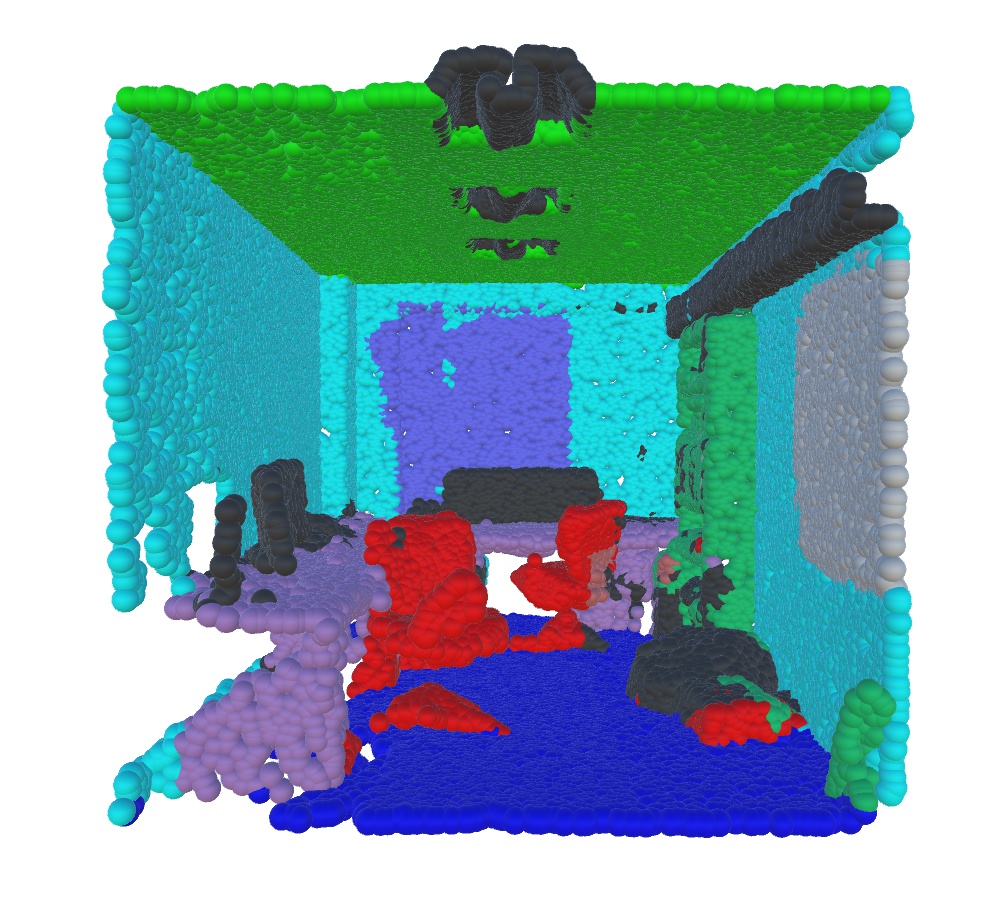}
  \includegraphics[width=0.22\textwidth]{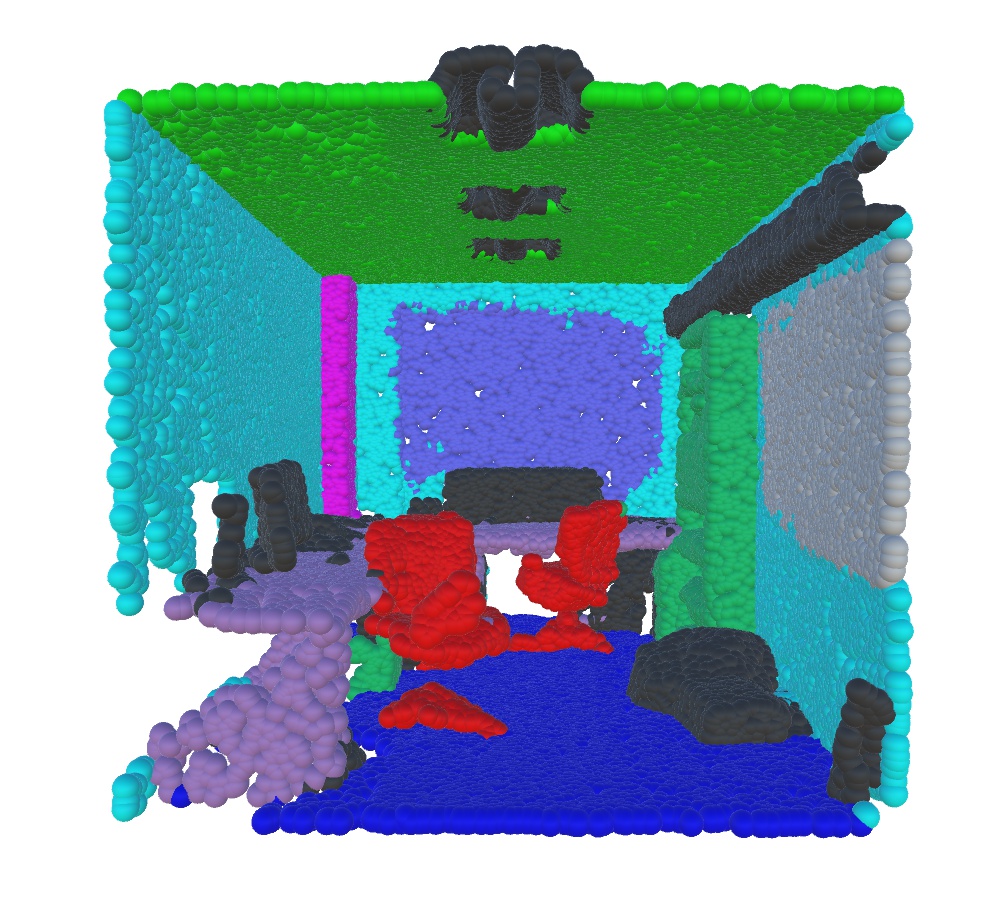}
  \includegraphics[width=0.22\textwidth]{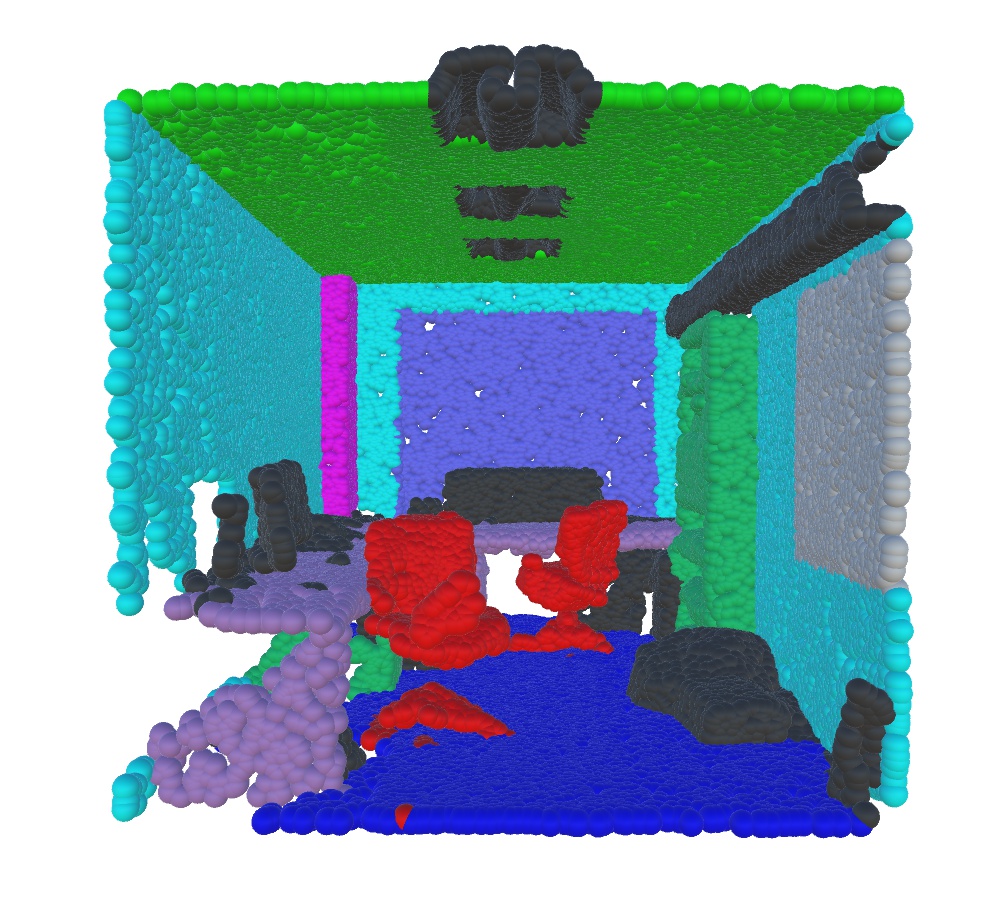}
  \includegraphics[width=0.22\textwidth]{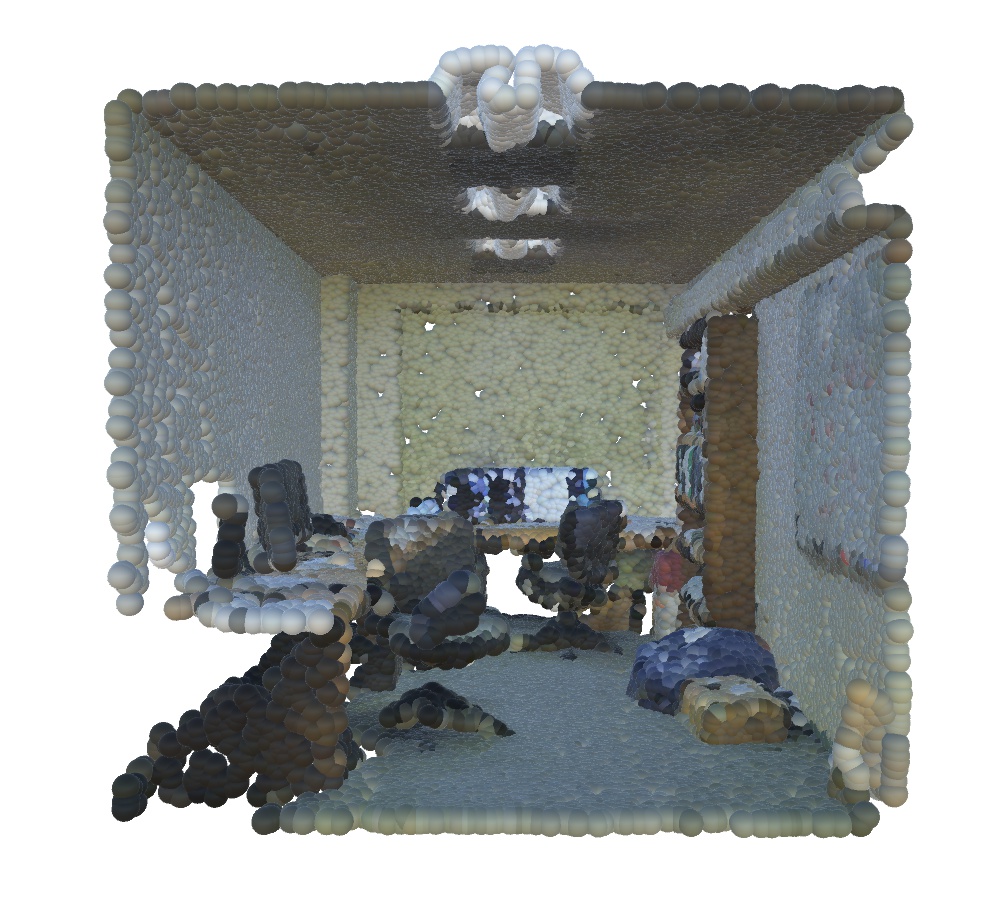} 
  \\

 \hspace{0.01\textwidth} {PointNet} \hspace{0.17\textwidth}  {Ours} \hspace{0.15\textwidth}  {Ground truth} \hspace{0.14\textwidth}  {Real color} 
  
  \caption{Semantic segmentation results. From left to right: PointNet, ours, ground truth and point cloud with original color. Notice our model outputs smoother segmentation results, for example, wall (cyan) in top two rows, chairs (red) and columns (magenta) in bottom two rows.}
  \label{fig:sem_seg}
\end{figure*}

\subsection{Indoor Scene Segmentation}
\label{sec:sem_seg}
\paragraph{Data} 
We evaluate our model on Stanford Large-Scale 3D Indoor Spaces Dataset (S3DIS) \cite{armeni_cvpr16} for a semantic scene segmentation task.
This dataset includes 3D scan point clouds for 6 indoor areas including 272 rooms in total. 
Each point belongs to one of 13 semantic categories---e.g.\ board, bookcase, chair, ceiling, and beam---plus clutter.
We follow the same setting as \citet{DBLP:journals/corr/QiSMG16}, where each room is split into blocks with area $1m \times 1m$, and each point is represented as a 9D vector (XYZ, RGB, and normalized spatial coordinates).
4,096 points are sampled for each block during training process, and all points are used for testing.
We also use the same 6-fold cross validation over the 6 areas, and the average evaluation results are reported.

The model used for this task is similar to part segmentation model, except that a probability distribution over semantic object classes is generated for each input point and no categorical vector is used here. 
We compare our model with both PointNet \cite{DBLP:journals/corr/QiSMG16} and PointNet baseline, where additional point features (local point density, local curvature and normal) are used to construct handcrafted features and then fed to an MLP classifier. 
We further compare our work with \cite{engelmann2017exploring} and PointCNN \cite{pointcnn}. \citet{engelmann2017exploring} present network architectures to enlarge the receptive field over the 3D scene. 
Two different approaches are proposed in their work: MS+CU for multi-scale block features with consolidation units; G+RCU for the grid-blocks with recurrent consolidation Units. 
We report evaluation results in Table \ref{table:sem_seg}, and visually compare the results of PointNet and our model in Figure~\ref{fig:sem_seg}.

\begin{table}[h]
\vskip 0.1in
\begin{center}
\resizebox{\columnwidth}{!}{
\begin{small}
\begin{sc}
\begin{tabular}{lccr}
\toprule
   &\quad Mean \quad   & \quad overall \quad \\
   & \quad IoU \quad   & \quad accuracy \quad \\
\midrule
PointNet (baseline) \cite{DBLP:journals/corr/QiSMG16}    & \quad 20.1 & 53.2 \\
PointNet \cite{DBLP:journals/corr/QiSMG16}                   & \quad 47.6 & 78.5 \\
MS + CU(2) \cite{engelmann2017exploring}                    & \quad 47.8 & 79.2  \\
G + RCU \cite{engelmann2017exploring}                        & \quad 49.7 & 81.1  \\
PointCNN \cite{pointcnn} 
                  & \quad \textbf{65.39} & - \\
\midrule
Ours                         &\quad 56.1 &  \textbf{84.1} \\
\bottomrule
\end{tabular}
\end{sc}
\end{small}
}
\end{center}
% \vskip -0.1in
\caption{3D semantic segmentation results on S3DIS. MS+CU for multi-scale block features with consolidation units; G+RCU for the grid-blocks with recurrent consolidation Units.}
\label{table:sem_seg}
\end{table}

% \subsection{Surface Normal Prediction } 
% \label{sec:normal_pred}
% We can also adapt our segmentation model to predict surface normals from point clouds. 

% \paragraph{Data} We still use the ModelNet40 dataset. The surface normals are sampled directly from CAD models. The normal of one point is represented by $(n_x, n_y, n_z)$. We  use 9,843 models for training and 2,468 models for testing. 

% \paragraph{Architecture} We change the last layer of our segmentation model to output 3 continuous values; mean squared error (MSE) is used as the training loss. 

% \paragraph{Results} Qualitative results are shown in Figure~\ref{fig:normal_est}, compared with ground truth. Our model faithfully captures orientation even in the presence of fairly sharp features.

%% file: sections/conclusion.tex
% !TEX root = ../tog.tex

\section{Discussion}
In this work we propose a new operator for learning on point cloud and show its performance on various tasks. Our model suggests that local geometric features are important to 3D recognition tasks, even after introducing machinery from deep learning. 

While our architectures easily can be incorporated as-is into existing pipelines for point cloud-based graphics, learning, and vision, our experiments also indicate several avenues for future research and extension. Some details of our implementation could be revised and/or re-engineered to improve efficiency or scalability, e.g.\ incorporating fast data structures rather than computing pairwise distances to evaluate $k$-nearest neighbors queries. We also could consider higher-order relationships between larger tuples of points, rather than considering them pairwise. Another possible extension is to design a non-shared transformer network that works on each local patch differently, adding flexibility to our model.

Our experiments suggest that intrinsic features can be equally valuable if not more valuable than point coordinates; developing a practical and theoretically-justified framework for balancing intrinsic and extrinsic considerations in a learning pipeline will require insight from theory and practice in geometry processing. Given this, we will consider applications of our techniques to more abstract point clouds coming from applications like document retrieval and image processing rather than 3D geometry; beyond broadening the applicability of our technique, these experiments will provide insight into the role of geometry in abstract data processing.

%% file: sections/acknowledgement.tex
% !TEX root = ../tog.tex
\begin{acks}
The authors acknowledge the generous support of Army Research Office grant W911NF-12-R-0011, of Air Force Office of Scientific Research award FA9550-19-1-0319, of National Science Foundation grant IIS-1838071, of ERC Consolidator grant No. 724228 (LEMAN), from an Amazon Research Award,  from the MIT-IBM Watson AI Laboratory, from the Toyota-CSAIL Joint Research Center, from the Skoltech-MIT Next Generation Program, and from Google Faculty Research Award. Any opinions, findings, and conclusions or recommendations expressed in this material are those of the authors and do not necessarily reflect the views of these organizations.
\end{acks}